\documentclass[journal]{vgtc}                     

\onlineid{1643}

\vgtccategory{Research}

\vgtcpapertype{algorithm/technique}

\title{RelMap: Reliable Spatiotemporal Sensor Data Visualization via Imputative Spatial Interpolation}

\author{%
  \authororcid{Juntong Chen}{0000-0001-9343-4032},
  \authororcid{Huayuan Ye}{0009-0008-8208-2017},
  \authororcid{He Zhu}{0009-0006-0835-7153},
  \authororcid{Siwei Fu}{0000-0001-8329-2448},
  \authororcid{Changbo Wang}{0000-0001-8940-6418},
  \authororcid{Chenhui Li}{0000-0001-9835-2650}
}

\authorfooter{
  \item J. Chen, H. Ye, H. Zhu, C. Wang, and C. Li are with the School of Computer Science and Technology, East China Normal University. Email: jtchen@stu.ecnu.edu.cn, huayuan221@gmail.com, 10215102469@stu.ecnu.edu.cn, \{cbwang, chli\}@cs.ecnu.edu.cn.
  \item S. Fu is with the School of Management, Zhejiang University. Email: fusiwei339@gmail.com.
  \item Chenhui Li is the corresponding author.
}

\abstract{%
Accurate and reliable visualization of spatiotemporal sensor data such as environmental parameters and meteorological conditions is crucial for informed decision-making. Traditional spatial interpolation methods, however, often fall short of producing reliable interpolation results due to the limited and irregular sensor coverage. This paper introduces a novel spatial interpolation pipeline that achieves reliable interpolation results and produces a novel heatmap representation with uncertainty information encoded. We leverage imputation reference data from Graph Neural Networks (GNNs) to enhance visualization reliability and temporal resolution. By integrating Principal Neighborhood Aggregation (PNA) and Geographical Positional Encoding (GPE), our model effectively learns the spatiotemporal dependencies. Furthermore, we propose an extrinsic, static visualization technique for interpolation-based heatmaps that effectively communicates the uncertainties arising from various sources in the interpolated map. Through a set of use cases, extensive evaluations on real-world datasets, and user studies, we demonstrate our model's superior performance for data imputation, the improvements to the interpolant with reference data, and the effectiveness of our visualization design in communicating uncertainties.
}

\keywords{Spatial interpolation, spatiotemporal data, uncertainty visualization, graph neural network}

\usepackage{amsfonts}
\usepackage{amsmath}
\usepackage{amssymb}
\usepackage{multirow}
\usepackage{siunitx}

\usepackage{graphicx}
\usepackage[normalem]{ulem}
\useunder{\uline}{\ul}{}

\usepackage[left=1.31cm,top=1.7cm,bottom=1.6cm,right=1.31cm]{geometry}

\AtBeginDocument{}

\def\*#1{\mathbf{#1}}
\def\cal#1{\mathcal{#1}}

\def\rm#1{\mathrm{#1}}

\newcommand\ttt[1]{\texttt{#1}}

\newcommand{\cjt}[1]{#1}
\newenvironment{cjte}{}{}

\newcommand{\bsc}[1]{%
  \tikz[baseline=-\the\dimexpr\fontdimen22\textfont2\relax]{
  \node[shape=circle, fill=black, draw, inner sep=0.5pt] (char) {
      \textcolor{white}{\small\sffamily\bfseries#1}};
  }%
}

\newcommand{\bs}[1]{%
  {\sffamily\bfseries#1}%
}

\DeclareDocumentCommand{\todo}{o}{
  \textcolor{red}{
    TODO\IfValueT{#1}{: #1}
  }
}

\usepackage{marginnote}
\usepackage{adjustbox}

\global\marginparsep=3pt

\newcommand{\sidecomment}[1]{%
  \ifdefined\revise%
  \marginnote{%
    \textcolor{Magenta}{%
      \adjustbox{minipage=0.35\marginparwidth,fbox}{%
          \scriptsize#1%
      }%
    }%
  }%
  \fi%
}%

\newcommand{\revision}[1]{%
  \ifdefined\revise%
  {\hypersetup{allcolors=Magenta}\color{Magenta}#1}%
  \else%
  #1%
  \fi%
}%

\teaser{
  \centering
  \includegraphics[width=0.88\linewidth]{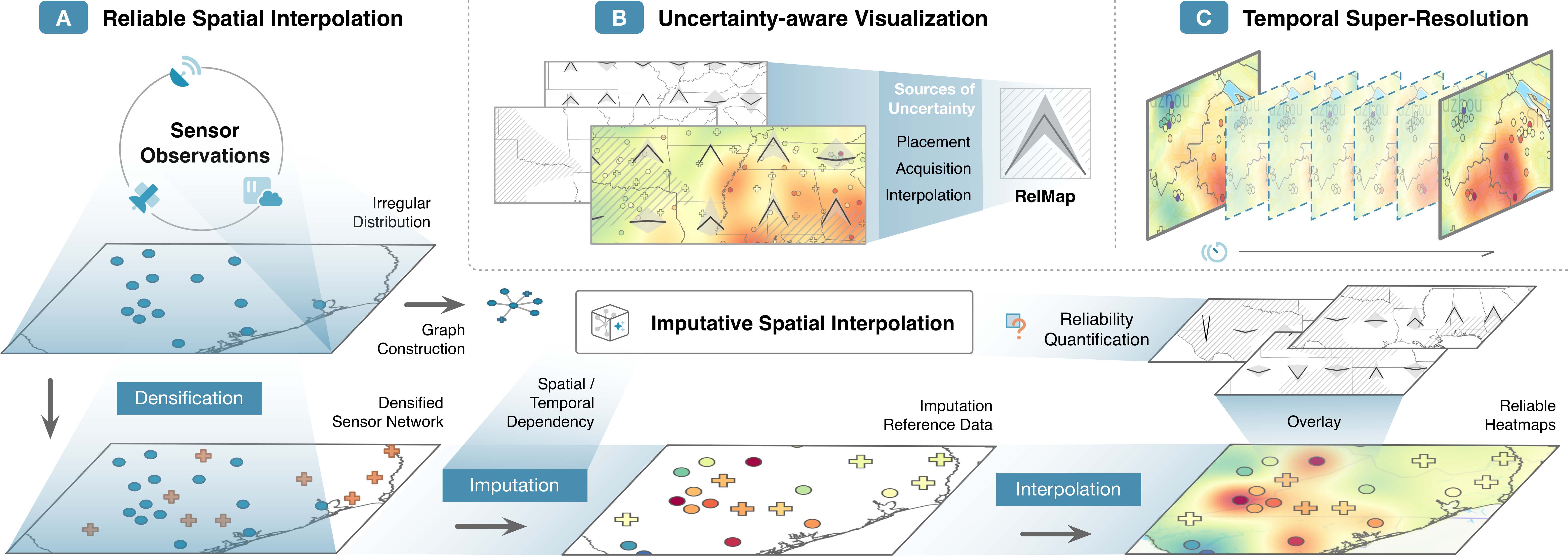}
  \caption{
We propose a novel spatiotemporal visualization pipeline leveraging imputation reference data produced by GNNs to generate more reliable and accurate heatmaps (\bs{A}). Various sources of uncertainties in the data-to-mapping process are visualized(\bs{B}). Additionally, our model can also increase the temporal resolution of the data, producing heatmaps at a finer granularity (\bs{C}).
  }
  \label{fig:teaser}
}

\graphicspath{{figs/}{figures/}{pictures/}{images/}{./}} 

\usepackage{tabu}                      
\usepackage{booktabs}                  
\usepackage{lipsum}                    
\usepackage{mwe}                       

\usepackage{mathptmx}                  

\begin{document}




\firstsection{Introduction}
\maketitle

Accurate and reliable visualization of spatiotemporal sensor data is crucial across various domains, such as environmental monitoring, urban planning, noise management, meteorological forecasting, etc. \cite{dengSurveyUrbanVisual2023}. Such data, typically collected from field sensors distributed across regions of interest, serves as the backbone for informed decision-making processes that can have significant socioeconomic impacts \cite{cheongEvaluatingImpactVisualization2016, chen2024salientime}. Heatmaps, which provide a continuous raster-based representation across the spatial domain, is a particularly widespread technique for visualizing these data \cite{liStreamMapSmoothDynamic2018}. Nowadays, billions of people rely on heatmaps to make their daily decisions, such as planning outdoor activities using precipitation and air quality maps \cite{prestonCommunicatingUncertaintyRisk2023}, selecting the most congestion-free route for travelling\cite{fengTopologyDensityMap2021}, or assessing the investment potential for real estate properties \cite{chenSenseMapUrbanPerformance2023}.

\revision{
In the context of visualizing spatiotemporal sensor data with heatmaps,
\sidecomment{R3.1\\R1.1}
three steps are involved: 1) data acquisition that involves collecting sensor readings, 2) spatial interpolation that transforms discrete sensor data into a continuous raster representation, and 3) visualization that presents the raster data to the audience.
Various challenges exist and we identify two research gaps.}
First, the distribution of sensors is often irregular, typically concentrated in urban areas because of physical constraints or budgetary considerations \cite{liSpatialInterpolationMethods2014}. Estimating values for areas with sparse sensor coverage is critical to reflecting the real scenarios. Yet, the task is challenging due to insufficient observations. Existing methods are built upon Tobler's first law of geography, which states that ``everything is related to everything else, but near things are more related than distant things'' \cite{toblerComputerMovieSimulating1970}. Hence, methods such as Inverse Distance Weighting (IDW) and Kriging estimate values based on the distance from the estimation center to nearby sensors \cite{oliverKrigingMethodInterpolation1990}, which may fail in areas that are far from urban areas.
Recent advances in Graph Neural Networks (GNNs) have demonstrated the capability to model spatial and temporal dependencies, and have been successfully applied in tasks such as traffic forecasting \cite{yuSpatioTemporalGraphConvolutional2018} and spatiotemporal data imputation \cite{wuInductiveGraphNeural2020}. However, the application in enhancing the reliability of spatial interpolation has not been fully explored.

Second, various uncertainties are introduced during the \cjt{data-to-mapping pipeline \cite{maceachrenVisualizingGeospatialInformation2005}, which refers to the entire process of sensor data acquisition, transformation, to visualization.} For example, sensors may produce erroneous readings due to sensor malfunctions, leading to extensive inaccurate areas in the heatmap \cite{liSpatialInterpolationMethods2014};
Uncertainty visualization in a heatmap is crucial for risk communication \cite{nowakDesigningAmbiguityVisual2023}, and may help users in decision-making\cite{joslynCommunicatingForecastUncertainty2010}. However, the majority of heatmaps employ colormaps to present density estimation\cite{retchlessGuidanceRepresentingUncertainty2016}, overlooking uncertainty in the visual representation.
Existing studies address this issue primarily by designing domain-specific visual analytics systems that reveal the uncertainty information through multi-coordinated views and interactions \cite{caoVoilaVisualAnomaly2018,zhangUncertaintyOrientedEnsembleData2021,nowakDesigningAmbiguityVisual2023}. However, adapting these approaches to real-world scenarios requires significant engineering workload and user learning effort.

\revision{
\sidecomment{R3.1\\R1.1}
In this paper, we propose \textit{RelMap}, a novel spatiotemporal data visualization pipeline that
leverages imputation reference data from Graph Neural Networks (GNNs),
generating reliable heatmaps results for spatiotemporal sample data with uncertainty information encoded.
}
\cjt{%
Our pipeline consists of three components: 1) Adaptive Sensor Densification (\cref{fig:pipeline}.a), which identifies optimal locations to place virtual sensors to fill in the reference data. The sampling process is based on existing sensors, with the goal of making the densified sensors uniformly distributed in sparse areas to improve spatial interpolation accuracy. 2) Imputative Interpolation (\cref{fig:pipeline}.b), which imputes values for the densified sensors using a Graph Neural Network (GNN) model. Our model incorporates Principal Neighborhood Aggregation (PNA) \cite{corsopna} and Geographical Positional Encoding (GPE) \cite{klemmer2023positional}, and effectively learns the spatiotemporal dependencies, providing additional benefit of temporal super-resolution and generates heatmaps at a finer granularity. The imputed values and original values are then combined to generate heatmap. 3) Uncertainty Visualization (\cref{fig:pipeline}.c), which use a novel visualization design to effectively communicates various sources of uncertainties. Our extrinsic, static design can be applied to any existing heatmaps, providing a general uncertainty visualization solution for interpolation-based heatmaps. Through extensive quantitative evaluations on real-world datasets and user studies, we demonstrate our model's superior data imputation capability, the efficacy of our pipeline for generating reliable heatmaps, and the effectiveness of our design in visualizing uncertainties.
}

In summary, our contributions are as follows:

\begin{itemize}[noitemsep,topsep=1pt]
    \item \revision{\sidecomment{R3.1\\R1.1}%
We introduce a novel visualization pipeline that leverages imputation reference data on the densified sensor network to reliably visualize spatiotemporal sample data in the form of heatmaps and quantifying geospatial uncertainties.

    \item We propose RelMap, an extrinsic, static visualization technique for interpolation-based heatmaps that effectively communicate various sources of uncertainties in the interpolation-based heatmaps.
    }
    \item We conduct extensive evaluations on real-world datasets to demonstrate the validity of our approach for data imputation and spatial interpolation, and conduct a user study to evaluate the effectiveness of our visualization design in communicating uncertainties.
\end{itemize}

\section{Related Work}

\subsection{Spatial Interpolation}

Spatial interpolation is the process of estimating values of unknown locations using observed data at known locations \cite{lamSpatialInterpolationMethods1983}. Spatial phenomena and environmental variables such as temperature, precipitation, and air quality are typically collected by point sampling as many areas are difficult to access, such as mountains or marine areas \cite{liSpatialInterpolationMethods2014}, necessitating spatial interpolation for producing continuous representations. As a fundamental procedure in geostatistics, various methods are proposed, which can be categorized into deterministic and probabilistic methods \cite{myersSpatialInterpolationOverview1994}. Deterministic methods, also referred to as non-geostatistical methods \cite{laslettComparisonSeveralSpatial1987}, such as Kernel Estimation, Inverse Distance Weighting (IDW), Triangulated Irregular Networks (TIN), and Radial Basis Functions (RBF) Interpolation, model interpolated surfaces from observed data points based on the distance to the target location.
Probabilistic methods, also referred to as geostatistical methods \cite{mitasSpatialInterpolation1999}, assume that the spatial variations can be modeled by a random process with spatial autocorrelation, thus requiring explicit modeling of autocorrelations \cite{matheronPrinciplesGeostatistics1963}. The most common geostatistical method is various Kriging variants that rely on different assumptions. Univariate methods such as Simple Kriging (SK), Ordinary Kriging (OK), and Cokriging characterize the spatial autocorrelation using semivariogram cloud, that is, the relationship between the semi-variance $\gamma(h)$ and the distance $h$ between pairs of points \cite{oliverKrigingMethodInterpolation1990}, and estimate values using the fitted model. Multivariate methods such as Universal Kriging (UK) and Regression Kriging (RK) incorporate auxiliary variables to make estimations \cite{henglRegressionkrigingEquationsCase2007}.

Due to the complexities and heterogeneity within geographical data that traditional statistical methods fail to capture \cite{zhuInferringSpatialInteraction2018}, recent studies
have applied deep learning techniques for spatial interpolation, either performed on the image domain to output the interpolation results in an end-to-end manner \cite{yanHighAccuracyInterpolation2021,zhuSpatialInterpolationUsing2020,zhouVoidsFillingMultiattention2022} or learn the spatial correlations from auxiliary data and output value distributions \cite{kirkwoodBayesianDeepLearning2022,zhangIntegrationMachineLearning2021,yaoSpatiotemporalInterpolationUsing2023}.

\subsection{Graph Neural Network for Data Imputation}

Data imputation refers to filling in missing values for observed locations \cite{miaoExperimentalSurveyMissing2023}. Several studies have applied deep learning techniques for data imputation tasks \cite{smiejaProcessingMissingData2018, applebyKrigingConvolutionalNetworks2020}.
In the context of spatiotemporal data, Spatiotemporal Graph Neural Networks have been proven to be an effective approach to learning local dependencies \cite{wuComprehensiveSurveyGraph2021}, \revision{\sidecomment{R2.6\\R3.2}%
with remarkable performance in various tasks such as traffic data forecasting \cite{yuSpatioTemporalGraphConvolutional2018,liDiffusionConvolutionalRecurrent2018,xu2020STMFGCN}, environmental sensor data recovery \cite{chen2022adaptiveAGCIN}, and spatiotemporal data imputation \cite{wuInductiveGraphNeural2020,yoonGAINMissingData2018}.
}

The foundation for most GNNs is message propagation,
a refinement of the convolution operation in CNNs,
where each node aggregates information from its neighbors and updates its embeddings \cite{wuComprehensiveSurveyGraph2021}. These GNNs are referred to as Convolutional Graph Neural Networks (ConvGNN).
Based on different message propagation mechanisms, ConvGNNs are categorized into Spectral and Spatial. Spectral-based methods utilize graph Laplacian matrix as a mathematical representation for graphs and perform message passing in the Fourier domain \cite{wuComprehensiveSurveyGraph2021}, such as SpectralCNN \cite{brunaSpectralNetworksLocally2014} and ChebNet \cite{defferrardConvolutionalNeuralNetworks2016}. Spatial-based methods directly operate on the graph structure and perform convolution in the spatial domain, such as Diffusion Graph Convolution \cite{atwoodDiffusionConvolutionalNeuralNetworks}, GraphSAGE \cite{hamiltonInductiveRepresentationLearning2017}, and Graph Attention \cite{velickovicGraphAttentionNetworks2018}.
\revision{\sidecomment{R2.6\\R3.2}%
To incorporate temporal information into GNNs, several studies have utilized RNN,
passing hidden states to a recurrent unit using graph convolutions \cite{liDiffusionConvolutionalRecurrent2018,jainStructuralRNNDeepLearning2016,cini2022fillingGRIN}. However, these techniques suffer from large computational overhead and gradient explosion problems \cite{wuComprehensiveSurveyGraph2021}.
CNN-based approaches, on the other hand, model temporal dependencies in a non-recursive manner and perform convolutions on the temporal dimension \cite{wuSpatialAggregationTemporal2021, yanSpatialTemporalGraph2018}.%
}

In our study, we propose a spatiotemporal graph neural network model using Principal Neighbourhood Aggregation (PNA) \cite{corsopna}, a spatial-based method that incorporates multiple aggregation functions for message passing. Combined with geographical positional encoding (GPE) \cite{klemmer2023positional}, our model can effectively learn the spatial and temporal dependencies for spatiotemporal data interpolation.

\subsection{Visualizing Geospatial Uncertainties}

\begin{figure*}[tbhp]
  \centering
  \includegraphics[width=0.9\textwidth]{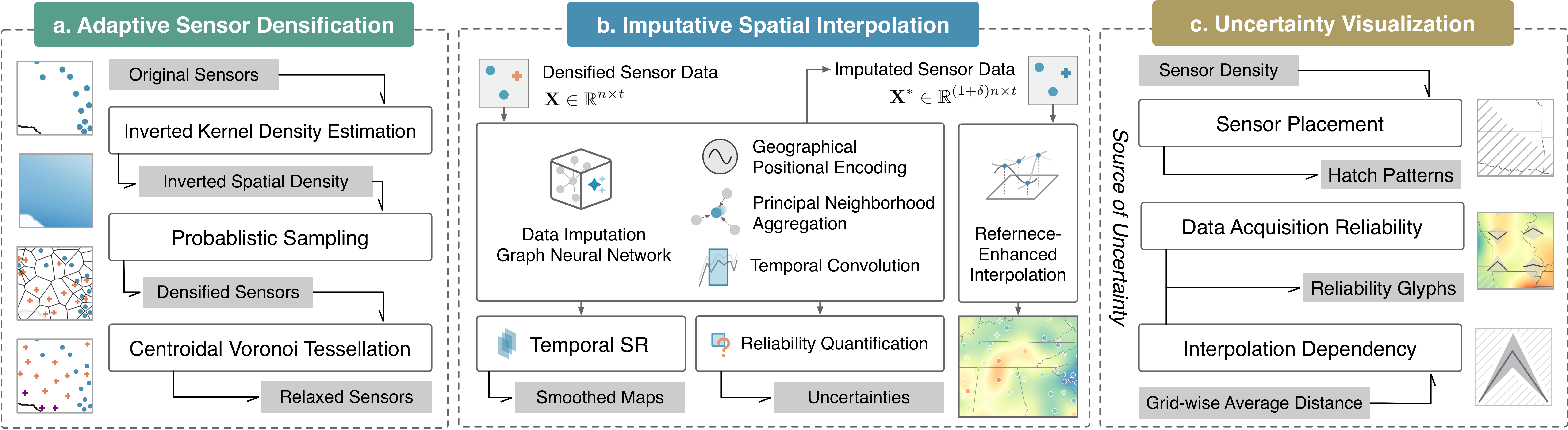}
  \caption{The overview of our pipeline. \bs{a}. Adaptive Sensor Densification: Identifying the optimal locations to fill in imputation reference data. \bs{b}. Imputative Spatial Interpolation: Leveraging GNNs to perform data imputation. The model is also used for quantifying the data acquisition reliability and temporal super-resolution.
  \bs{c}. Uncertainty Visualization: Communicate various sources of uncertainties in interpolation-based heatmaps.}
  \label{fig:pipeline}
\end{figure*}

\revision{
Visualization of geospatial uncertainty has been an active field of research in both geoscience and visualization \cite{maceachrenVisualizingGeospatialInformation2005,hullmanPursuitErrorSurvey2019}. These uncertainties are widespread and stem from a variety of sources. Pang et al. \cite{pangApproachesUncertaintyVisualization1997} identified three types of uncertainty:
collection uncertainty, 
derived uncertainty, 
and visualization uncertainty. 
Effectively visualizing the uncertainties is crucial for communicating risks and enhancing users' confidence \cite{sarmaEvaluatingUseUncertainty2023} and information retrieval performance \cite{korporaalEffectsUncertaintyVisualization2020} when making decisions.
}

Visualizing geospatial uncertainties is particularly challenging as they often need to be integrated with the underlying thematic data \cite{retchlessGuidanceRepresentingUncertainty2016}. Prior research \cite{brewerDesigningBetterMaps2015, retchlessGuidanceRepresentingUncertainty2016, maceachrenVisualizingGeospatialInformation2005} has emphasized that visual representations for uncertainties should complement, rather than hinder map-reading tasks and facilitate decision-making. The commonly used distinction for map-based uncertainty visualizations is intrinsic and extrinsic \cite{gershonVisualizationImperfectWorld1998,kinkeldeyHowAssessVisual2014}. Intrinsic methods modify the existing symbology or color encodings, such as Risk Contours \cite{klippelInterpretingSpatialPatterns2011} that uses contour lines to highlight the percentiles of interpolation results, and Dotmaps \cite{prestonCommunicatingUncertaintyRisk2023, kayWhenIshMy2016} that rendering multiple potential values within a single pixel. Extrinsic methods introduce new elements or interactions to display uncertainty, such as glyphs, textures, or extra views. For instance, anomaly glyphs in Viola \cite{caoVoilaVisualAnomaly2018} reveal the spatial data anomaly likelihood using different saturation levels. Zhang et al. \cite{zhangUncertaintyOrientedEnsembleData2021} propose region stability heatmaps to display the stability of numerical simulation results. TPFlow \cite{liuTPFlowProgressivePartition2019} employs circle-shaped glyphs on maps to encode the deviation of the raw data from the regular patterns in the spatial domain. VFDP \cite{chenVFDPVisualAnalysis2022} utilizes different colors of textures to represent the variation of flight delay evolutions. These techniques are either dynamic or static, according to whether interactivity is involved \cite{kinkeldeyHowAssessVisual2014}. While dynamic techniques provide more flexibility and encoding ability, they require more complex implementation and have compromised applicability and accessibility compared with static techniques.

Inspired by these techniques and guidelines of map-based uncertainty visualization \cite{retchlessGuidanceRepresentingUncertainty2016}, we propose an extrinsic, static visualization design that effectively conveys uncertainties, capable of applying to any existing interpolation-based heatmaps.

\section{Methodology}

\subsection{Overview}

\cref{fig:pipeline} illustrates the overview of our RelMap pipeline. First, instead of performing interpolation directly on the original data, we use a densification, imputation, and interpolation process. The Adaptive Sensor Densification process (\cref{fig:pipeline}.a) identifies optimal locations to place virtual sensors, sampling locations with priorities in areas with sparse sensor coverage.
The value of these virtual sensors are first set to 0 and then imputed using a GNN-based model, where we integrate Principal Neighborhood Aggregation (PNA) and Geographical Positional Encoding (GPE) to capture the spatial and temporal dependencies. Subsequently, we employ an RBF interpolator on both the original and imputed sensor data to generate heatmaps (\cref{fig:pipeline}.b).

Despite this process, the interpolation-based heatmaps still contain various uncertainties. To convey this information, we propose a static, extrinsic visualization design named RelMap, encoding various sources of uncertainty including sensor placement, data acquisition, and spatial interpolation. Reliability of different locations is quantified using our model, where we generate a reference value for each sensor and quantify the reliability based on the deviation of observed values from the reference.

Formally, let the input spatiotemporal sample data be $\mathbf{X} \in \mathbb{R} ^ {n \times t}$, where $n$ is the total number of available sensors $S = \{ s_1, \dots s_n \}$ associated with their geolocations $\langle \mathrm{lng}, \mathrm{lat} \rangle$ and $t$ is the total number of time steps. The imputative spatial interpolation pipeline is formulated as
$I: (\mathcal{F}(\mathcal{G} (\*X, \Lambda ), N, \epsilon)) \rightarrow \*Y$, where $\mathbf{Y} \in \mathbb{R} ^ {h \times w \times t}$ is a continuous raster representation of sample data. Here $h$ and $w$ are the height and width of the output raster. The GNN-based data imputation model $\mathcal{G}: (\mathbf{X}, \Lambda) \rightarrow \mathbf{X}^* \in \mathbb{R} ^ {(1 + \delta)n \times t}$, takes masked sensor data as input and outputs imputed data ${\mathbf{X}^*}$ as additional references for spatial interpolation, where $\delta$ being the ratio of sensor densification. During the process, the hidden feature of the $l$-th layer of the GNN is denoted as $X^l$, and the output of the last layer is $X^*$. Then, using original sensor data and densified sensor data, we employ an RBF spatial interpolator, denoted as $\mathcal{F}: (\mathbf{X}^*, N, \epsilon) \rightarrow \mathbf{Y}$, where $\epsilon$ and $N$ are interpolation parameters controlling the shape of the kernel and the number of nearest neighbors, respectively.


\subsection{Adaptive Sensor Densification}

\cjt{
To identify optimal locations to fill in imputation reference data, we design an adaptive sensor densification module to add virtual sensors to the original sensor network.} The values for these virtual sensors, initially set to 0, are subsequently filled in by the model as described in \cref{sec:network}. Given a densification ratio $\delta$ and the original sensor sets $S$, we aim for the densified sensor set $S' = \{s'_1, \dots, s'_{\lfloor \delta n \rfloor} \}$ that fulfills two criteria: 1). The placement of virtual sensors should be prioritized in areas with sparse sensor coverage, and 2). The overall sensor distribution should be as uniform as possible, avoiding clustering from the original sensors.

We achieve this by performing Kernel Density Estimation on the original sensor locations to characterize their spatial distribution. Then, we use Centroidal Voronoi Tessellation (CVT) to relax the sensors sampled from the inverted density, ensuring their uniform distributions.

\begin{figure}[hbt]
  \centering
  \includegraphics[width=1\columnwidth]{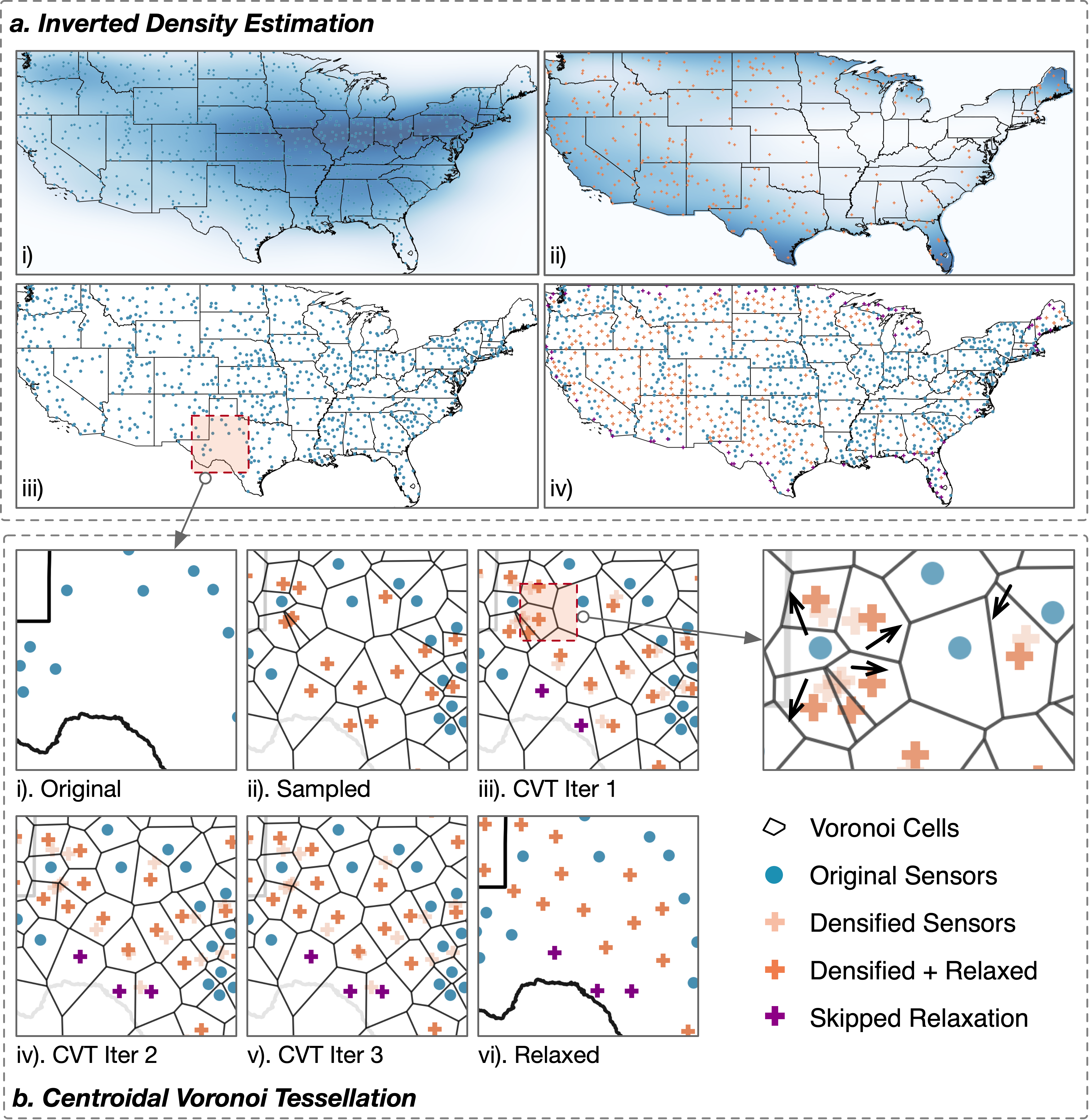}
  \caption{
    The process of Adaptive Sensor Densification. \bs{a}. The density and distribution for original sensors (i, iii), initially densified sensors (ii), and the entire sensor network after densification and relaxation. \bs{b}. The densified sensor locations after each CVT iteration. The densified sensors are moved to the centroid of their Voronoi cells in each iteration (iii, iv, v) to avoid clustering with the original sensors (vi). After 3 iterations, the densified sensors are reasonably distributed in this example.
  }
  \label{fig:cvt}
\end{figure}

\subsubsection{Inverted Density Estimation}
\label{sec:kde}

Kernel Density Estimation (KDE) is a widely used non-parametric technique for estimating the probabilistic density function of data. In our pipeline, we employ two-dimensional KDE on the coordinates of original sensors to obtain the spatial distribution $D(s)$, formulated as:

\begin{equation}
  D(s) = \frac{1}{n} \sum_{i=1}^{n} K_h\Big(\frac{\text{dist}(s - s_i)}{h}\Big),
\end{equation}

\noindent
where $K$ is a standard Gaussian kernel with bandwidth $h$, computed according to Silverman's rule of thumb \cite{silvermanDensityEstimizationStatistics1986}. To encourage densification in areas with sparse sensor coverage, we use the inverted density as the sampling probability:  $\bar{D}(s) = \max(e^{-\lambda D(s)} - \theta, 0)$,
where $\lambda$ is a scaling factor that amplifies the focus on sparsely populated areas, and $\theta$ acts as a density threshold. Sampling is restricted in areas with $\bar{D}(s) > \theta$. The densification probability $\bar{D}(s)$ is then normalized into the range of $[0, 1]$, and areas outside the terrain boundary are assigned to $0$. We proceed to probabilistically sample $\lfloor \delta n \rfloor$ sensors based on $\bar{D}$. \cref{fig:cvt}.a.i) and \cref{fig:cvt}.a.ii) shows the density and inverted density for sensors in the \ttt{ushcn} dataset.

\subsubsection{Centroidal Voronoi Tessellation}

The virtual sensors sampled on $\bar{D}$ may be clustered with original sensors, as shown in \cref{fig:cvt}.b.ii). To optimize their locations, we employ the Centroidal Voronoi Tessellation (CVT) algorithm. CVT, inspired by the Lloyd algorithm, is a relaxation-based technique for blue noise sampling, which is a type of noise with weak low-frequency components and uniform distribution of points in a given area \cite{yanSurveyBlueNoiseSampling2015}.

We first build a Voronoi diagram based on the original and densified sensors. The Voronoi diagram partitions the sampling domain into $n$ cells, where each cell is associated with a sensor and contains all points nearer to their corresponding sensor than any other sensor. Let the Voronoi cell associated with each sensor be $s_i$ be $V_i$, during each iteration, CVT moves the sensors to the location that minimizes the following energy function:
$
E_{\text{CVT}}(X) = \sum_{i=1}^{n} \int_{V_i} \lVert x - x_i \rVert^2 \, \rm{d} x .
$
The optimal target location is the centroid of the Voronoi cell. During each iteration, the relaxation is only applied to densified sensors. Cells extending beyond the terrain's boundaries are excluded since the target location may fall outside the sample domain. We rebuild the Voronoi diagram after each relaxation and repeat this process. \cref{fig:cvt}.b shows the relaxed sensors after 3 iterations. In practice, we found that 2 to 3 iterations are sufficient to relax the densified sensors to a reasonable distribution.

\subsection{Data Imputation Network}
\label{sec:network}

To estimate the values of the densified virtual sensors, we leverage a Graph Neural Network (GNN) model to impute their values.
We model the sensor network as a fully connected graph with time-varying values, where each sensor is a node and the edges represent spatial relationships.

\subsubsection{Network Architecture}

\begin{figure}[htbp]
  \centering
  \includegraphics[width=1\columnwidth]{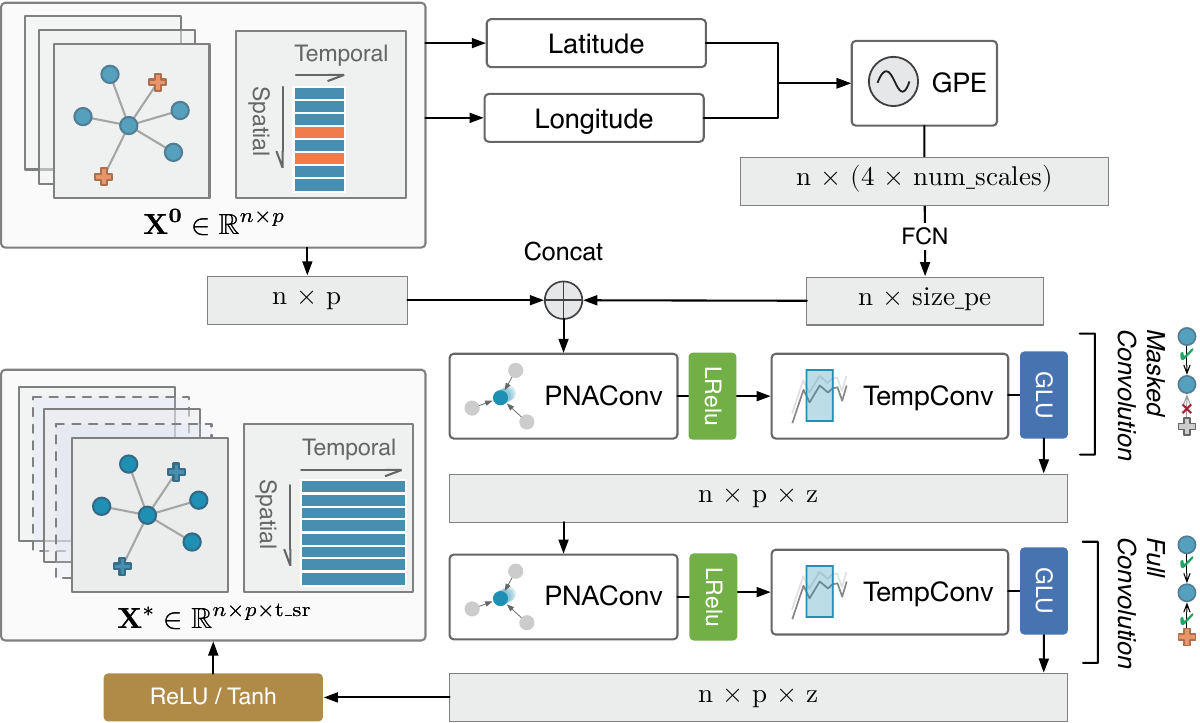}
  \caption{
    Network architecture. The first PNA layer uses masked convolution with $A_{\text{first}}$, passing messages only between original sensors. Later layers use full convolution with $A_{\text{sub}}$, allowing messages to pass between all sensors.
  }
  \label{fig:net}
\end{figure}

\Cref{fig:net} shows the network architecture, consisting of three building blocks: the Geographical Positional Encoding (GPE) module, Principal Neighborhood Aggregation (PNA) Spatial Convolution Module, and Temporal Convolution (TC) Module. The GPE module encodes the geographical information of the sensors, the PNA Spatial Convolution Module consists of multiple aggregators and serves as an effective message-passing mechanism to capture the spatial dependencies, and the Temporal Convolution Module learns the time-varying patterns of the sensors.

The sensor network and their values at time $t$ are modeled as a complete graph $G_t = (S, E)$, where $S$ is the sensor set $\{s_n\}_1^n$ and $E$ is the edge set. The model takes $p$ consecutive graphs $G_{\text{in}} = \{G_1 \dots G_p\}$ as input, represented as a value matrix $X^0 \in \mathbb{R}^{n \times p}$ and an adjacency matrix $A \in \mathbb{R}^{n \times n}$, where $n$ is the number of sensors, $p$ is the size of temporal window. The input contains both original $S_o$ and densified sensors $S_d$, where $X^0[s, :] = 0, \forall s \in S_d$. The adjacency matrix $A$ for the first PNA layer ($A_{\text{first}}$) and subsequent PNA layers ($A_{\text{sub}}$) differ. $A_{\rm{first}}$ limits message passing to $S_o$ and the top $k$ nearest neighbors, formulated as:

\begin{equation}
  A_{\text{first}}[i, j] = \begin{cases}
    e^{-H(s_i, s_j)} & , \text{ if } s_i, s_j \in S_o  \text{ and } s_j \in N_k(s_i) \\
    0                & , \text{ otherwise }
  \end{cases},
\end{equation}

\noindent
where $H(\cdot)$ is the Haversine distance, $N_k(s_i)$ denotes the top $k$ nearest neighbors of sensor $s_i$. For the subsequent PNA layers, the model allows message passing between all sensors, formulated as:

\begin{equation}
  A_{\text{sub}}[i, j] = \begin{cases}
    e^{-H(s_i, s_j)} & ,\text{ if } s_j \in N_k(s_i) \\
    0                & ,\text{ otherwise}
  \end{cases}.
\end{equation}

$A_{\text{first}}$ and $A_{\text{sub}}$ are then normalized into the range of $[0, 1]$. Before spatial convolution, the GPE module transforms the sensor coordinates into positional encodings, concatenating them with the data matrix $X$ to provide the model with both topological and geographical information about the sensor network. Subsequent layers alternate between spatial and temporal convolutions, followed by a 1D convolutional and output activations to obtain the recovered data $X^* \in \mathbb{R}^{n \times (p \times \text{t\_sr})}$, thus achieving data imputation and, if required, temporal super-resolution. We train the model using a masked subgraph training strategy, as detailed in \cref{sec:training}.

\subsubsection{Geographical Positional Encoding}

The adjacency matrix $A$, derived from sensor distances, can only represent the topological relations but omits their spatial contexts.
Previous studies such as GPS2Vec \cite{yinGPS2VecGeneratingWorldwide2019} and Space2Vec \cite{maiMultiScaleRepresentationLearning2020} have demonstrated the necessity of integrating geographical location into models for geospatial tasks.
In our model, we incorporate Geographical Positional Encoding (GPE) \cite{klemmer2023positional}, a technique inspired by transformers \cite{vaswaniAttentionAllYou2017} to explicitly map the sensor locations into a learnable latent space.

Let the coordinates of each sensor $s_i$ be $\langle \text{lng}_i, \text{lat}_i \rangle$.The coordinates are first normalized to delta coordinates within the range of $[-1, 1]$, where the average value is subtracted from all longitudes and latitudes. This yields a coordinate matrix $\mathbf{C} = [\mathbf{x}, \mathbf{y}]$ where $\mathbf{x} = [x_1, \dots x_n]^\intercal$ and $\mathbf{y} = [y_1, \dots y_n]^\intercal$. Applying a sinusoidal transform at scale $m$, we get:

\begin{equation}
  \mathbf{ST}_m = \Big[
    \cos(\frac{\mathbf{x}}{2^{m-1}}), \sin(\frac{\mathbf{x}}{2^{m-1}}), \cos(\frac{\mathbf{y}}{2^{m-1}}), \sin(\frac{\mathbf{y}}{2^{m-1}})
    \Big],
\end{equation}

\noindent
where $1 \le m \le M$ is the current scale. A total of $M$ scales are used to encode the coordinates, capturing the spatial context at different levels of detail. These transformations are concatenated and passed into a fully connected layer with learnable parameter $\Theta$ to obtain the geographical encoding matrix $\mathbf{PE} \in \mathbb{R}^{n \times 4M}$:

\begin{equation}
  \*{PE} = \rm{FC}(\begin{bmatrix}
    \rm{ST}_1, \dots, \rm{ST}_M
  \end{bmatrix}, \Theta),
\end{equation}

The geographical encoding matrix $\mathbf{PE}$ is concatenated with the input data $X^0$ to construct the input for the spatial convolution layer.

\subsubsection{Spatial Convolution}

Spatial Convolution is the most important component for GNNs, as it governs how the latent space embeddings propagate along the graph structure. During the forward pass, each node collects the information from its neighbors to update its embedding. Employing a single aggregator like \ttt{MEAN} or \ttt{MAX} may fail to differentiate between neighborhood messages thereby compromising model performance \cite{corsopna}. In our model, we employ Principal Neighborhood Aggregation (PNA) \cite{wuSpatialAggregationTemporal2021} as the spatial convolution module, which is an expressive message-passing mechanism that uses multiple aggregators and scalers.

\begin{figure}[htbp]
  \centering
  \includegraphics[width=1\columnwidth]{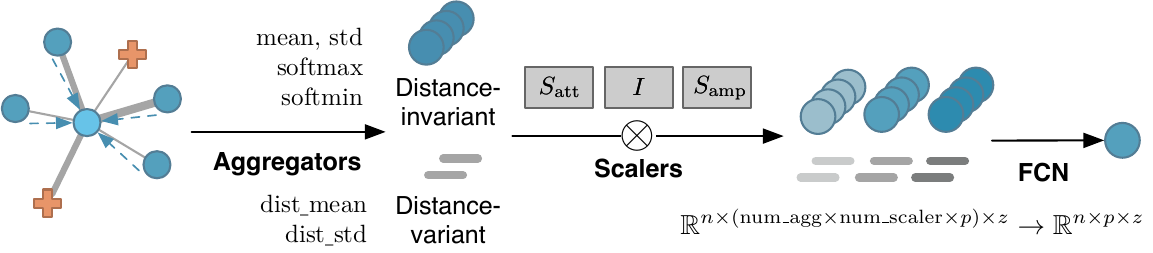}
  \caption{
    PNA convolution process. The aggregators and scalers are combined using the tensor product, and the result is passed through a fully connected layer to obtain the output of the spatial convolution.
  }
  \label{fig:pna}
\end{figure}

In our model, we consider two types of aggregators to collect messages from node neighbors: distance-invariant and distance-variant. Distance-invariant aggregators including $F_{\text{mean}}$, $F_{\text{softmax}}$, $F_{\text{softmin}}$, and $F_{\text{std}}$ consider node values and aggregate using different operations, as proposed by Corso et el. \cite{corsopna}. Let $X^l_{[i,:]}$ be the feature of node $i$ for the $l$-th layer, the distance-invariant aggregators are defined as:

\begin{equation}
  F(X^l_{[i,:]}) = \sigma(\{X^l_{[j,:]}\}_1^{|\Omega_i|}), j \in \Omega_i,
\end{equation}

\noindent
where $\sigma(\cdot)$ is the specific aggregation function \texttt{MEAN}, \texttt{SOFTMAX}, \texttt{SOFTMIN}, and \texttt{STD}; $\Omega_i$ denotes the set of neighborhood node $i$ that satisfies $A[i, j] > 0$.
Distance-variant aggregators, on the other hand, take into account the distances between sensors to characterize the spatial context of the sensors. Let $A^l$ be the adjacency matrix of the $l$-th layer, the distance variant aggregators $F_{\text{dmean}}$ and $F_{\text{dstd}}$ are defined as:

\begin{align}
  F_{\text{dmean}}(X^l_{[i,:]}, A^l) & = \frac{1}{|\Omega_i|}\sideset{}{_{j \in \Omega_i}}\sum A^l_{[j, i]} \\
  F_{\text{dstd}}(X^l_{[i,:]}, A^l)  & = \sqrt{
    \text{ReLU} \Big[
      F_{\text{dmean}} \big( (A^l_{[i:,]})^2 \big) - \big(F_{\text{dmean}}(A^l_{[i:,]})\big)^2
      \Big] + \epsilon
  },
\end{align}

\noindent
where $\text{ReLU}$ is the Rectiﬁed Linear Unit to avoid negative values and $\epsilon$ is a small positive number to ensure the differentiability. With the above aggregators, the target node can leverage the average, extreme values, and the diversity of input signals when passing messages.

Additionally, we employ scalers to adjust the importance of different aggregators. The scalers are computed using the number of messages received by a node, thereby describing the neighborhood influence and avoiding small signals being ignored. For our network, we use the sum of the edge weights to denote the signal strength. The amplification scaler ($S_{\text{amp}}$) and attenuation scaler ($S_{\text{att}}$) are defined as:

\vspace{-2pt}
\begin{align}
  S_{\text{amp}}(d) & = \frac{\log(d + 1)}{\eta}, S_{\text{att}}(d) = (S_{\text{amp}})^{-1},
\end{align}
\vspace{-2pt}

\noindent
where $\eta$ is the average distance of all sensors. The PNA convolution then takes a tensor product, denoted as $\otimes$, to combine the scalers and aggregators, where they are multiplied and concatenated column-wise:

\begin{equation}
  \text{PNA} = \begin{bmatrix}
    I              \\
    S_{\text{att}} \\
    S_{\text{amp}}
  \end{bmatrix} \otimes
  \begin{bmatrix}
    F_{\text{mean}}(X^l)    \\
    \vdots \hspace{4pt} \text{\raisebox{2pt}{\small(all scalers)}} \\
    F_{\text{dstd}}(X^l, A^l)
  \end{bmatrix}
\end{equation}

Subsequently, we apply a fully connected layer parameterized by weights $\Theta$ and offset $b$ to obtain the next layer's node embedding $X^{l+1}$:

\begin{equation}
  X^{l+1}_{[i,:]} = \text{ReLU}\big(\Theta^l \text{PNA}(X^l_{[i,:]}, A^l_{[i,:]}) + b^l \big)
\end{equation}

\Cref{fig:pna} illustrate the spatial convolution process. Aggregators and scalers in PNA convolution can discriminate and capture various spatial distributions and data patterns, effectively enhancing the model's data imputation capabilities, as evaluated in \cref{sec:eval-imputation}.

\subsubsection{Temporal Convolution}

Temporal Convolution (TC) integrates time-series information into the model, enabling nodes to incorporate latent features in the temporal domain during message passing. Some studies have leveraged Recurrent Neural Networks (RNN) to model the temporal dependencies \cite{seoStructuredSequenceModeling2018,jainStructuralRNNDeepLearning2016}. In our model, we use the convolution operator since it is more parameter-efficient and has the flexibility to handle varying input lengths, which is crucial for processing time-series data in dynamic environments.

Let $X^l \in \mathbb{R}^{n \times p \times z}$ be the input feature at the $l$-th layer, where $p$ is the temporal window length, $z$ is hidden feature size for each sensor. TC is a 2D convolution over the temporal domain, followed by a Gated Linear Unit (GLU) activation \cite{dauphinLanguageModelingGated2017}:

\begin{equation}
  X^{l+1} = (W^l \ast X^l + b^l) \odot \sigma(W'^l \ast X^l + b^l),
\end{equation}

\noindent
where $\ast$ is the convolution operation,
$W^l, W'^l \in \mathbb{R}^{w \times 1}$
are the convolution kernels and $b^l, b'^l$ are the bias terms for GLU activation, $\sigma$ is a sigmoid activation applied to a gating portion of the convolution output. The GLU activation controls if the information should be passed to the next layer. We apply SAME padding to the convolutions to ensure the input and output sizes are the same. If temporal super-resolution is specified, we increase the number of kernels of the convolution to multiply the output size to $X^* \in \mathbb{R}^{n \times (p \times \text{t\_sr}) \times z}$.

\subsubsection{Training with Masked Subgraphs}
\label{sec:training}

We employ a masked subgraph training strategy to enable the model to capture spatial and temporal dependencies. This involves simulating the data imputation process by randomly sampling subgraphs with a portion of sensors masked, and supervising the model by minimizing the differences between its imputed data and the actual data.

\begin{figure}[htbp]
  \centering
  \includegraphics[width=1\columnwidth]{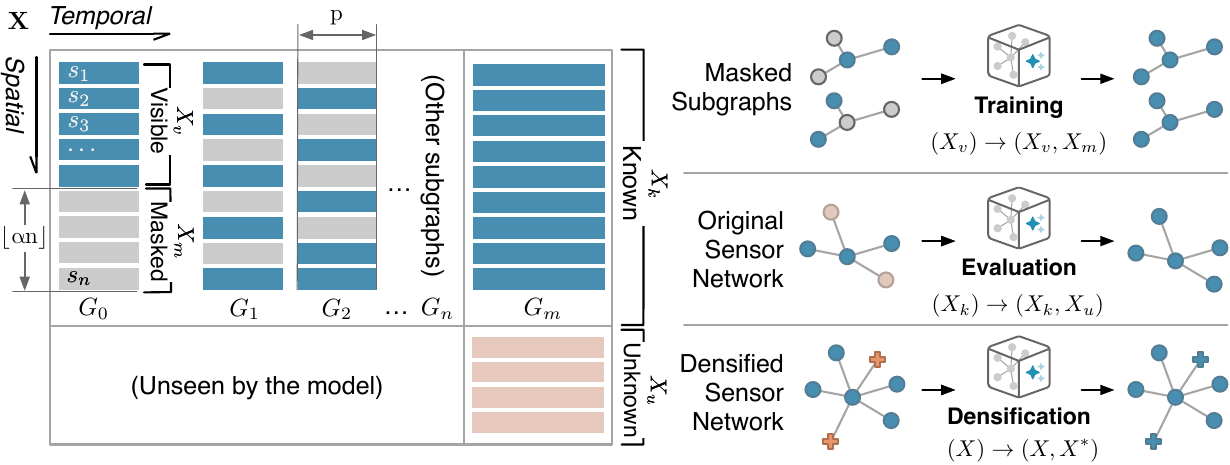}
  \caption{
    Training procedure. The model takes randomly sampled subgraphs with masked data as input and performs data imputation for the masked sensors. For evaluation, we use a different set of sensors ($X_m$) and compare the model's imputation with the ground truth.
  }
  \label{fig:train}
\end{figure}

\Cref{fig:train} illustrates the training procedure. We partition the data $X \in \mathbb{R}^{n \times p}$ across temporal and spatial dimensions into four subsets: masked $X_m$, visible $X_v$, known $X_k$, and unknown $X_u$. Let the masked rate be $\alpha$, the temporal window size be $p$. During each training epoch, we sample a subgraph $G_m$ from $X_k$ with duration $p$. The sensors and starting temporal positions are selected randomly. We then mask $\lfloor \alpha n \rfloor$ sensors by setting their values to 0 to obtain $X_m$ and $X_v$. The model takes $X_v$, $X_m = 0$, and adjacency matrix $A$ as input, recovering imputation values for $X_m$. The loss is computed against $X_k$. During the evaluation, we use $X_k$ and $X_u$ that are unseen by the model as input, comparing the model's imputation for $X_u$ against the truth. We employ Huber Loss \cite{huberRobustEstimationLocation1964}, a combination of L1 loss and L2 loss, as our loss function. L1 loss is more robust to outliers, while L2 loss is more sensitive to small errors and nuanced disturbance. Employing Huber loss allows the model to produce more accurate references for interpolation while avoiding overfitting to extremums or anomalies. The Huber loss is formulated as:

\begin{equation}
  \mathcal{L}_{\text{huber}} = \begin{cases}
    \frac{1}{2}(X - X^*)^2 & , \text{ if } |X^0 - X^*| \le \gamma  \\
    \gamma \cdot \big(|X - X^*| - \frac{1}{2}\gamma\big) & , \text{ otherwise },
  \end{cases}
\end{equation}

\noindent

where $\gamma$ is a hyperparameter controlling the threshold of using L1 loss or L2 loss. Our model is trained with the Adam \cite{kingmaAdamMethodStochastic2014} optimizer with a 0.05 dropout ratio applied. As a parameter-efficient model, the training process can be completed within hours using a single NVIDIA RTX 4070 GPU for datasets with thousands of sensors and time steps.

\subsection{Reference-Enhanced Spatial Interpolation}

With the imputation data filled in the virtual sensors by the model, we enhance the spatial interpolation by incorporating these reference values into the original sensor network. We employ Radial Basis Function (RBF) interpolation, a widely used interpolation method using irregularly distributed observations \cite{liSpatialInterpolationMethods2014}. For imputed sensor network $S = \{s_1, \cdots, c_n\}$ and their values $\{x_i, \cdots, c_n\}$, RBF interpolation is a combination of RBF functions and a polynomial of degree $q$, formulated as:

\vspace{-2pt}
\begin{equation}
  \*Y = \cal{F}(s) = \sum_{i=1}^{n}c_i \phi(\| s - s_i \|)  + \sum_{k=1}^{q} d_k p_k(s)
  ,
\end{equation}
\vspace{-2pt}

\noindent
where $\phi(\| \cdot  \|)$ is a univariate positive definite RBF function whose value decreases as the distance to the center increases. Common choices for $\phi$ include Gaussian, Thin-plate Spline, Multiquadric, etc. \cite{fasshauerChoosingOptimalShape2007}. $p_k(s)$ are monomials with a total degree of $q$. Weighting coefficients $\mathbf{c} = [c_1, \cdots, c_n ]^\intercal$
are solvable by the interpolation condition:

\vspace{-2pt}
\begin{equation}
  (\*D + \lambda I)\*c + \*P \*b = \*x ,
\end{equation}
\vspace{-2pt}
\noindent
where $\*D[i,j] = \phi(\| s_i - s_j \|)$, $\*x = [x_1, \cdots, x_n]^\intercal$, $\*P$ is the matrix of monomials evaluated at the sensor locations and $\*b = [b_1, \cdots, b_q]^\intercal$. $\lambda$ is a non-negative smoothing parameter, controlling the trade-off between the interpolation error and the smoothness of the interpolant.

\begin{figure}[htbp]
  \centering
  \includegraphics[width=1\columnwidth]{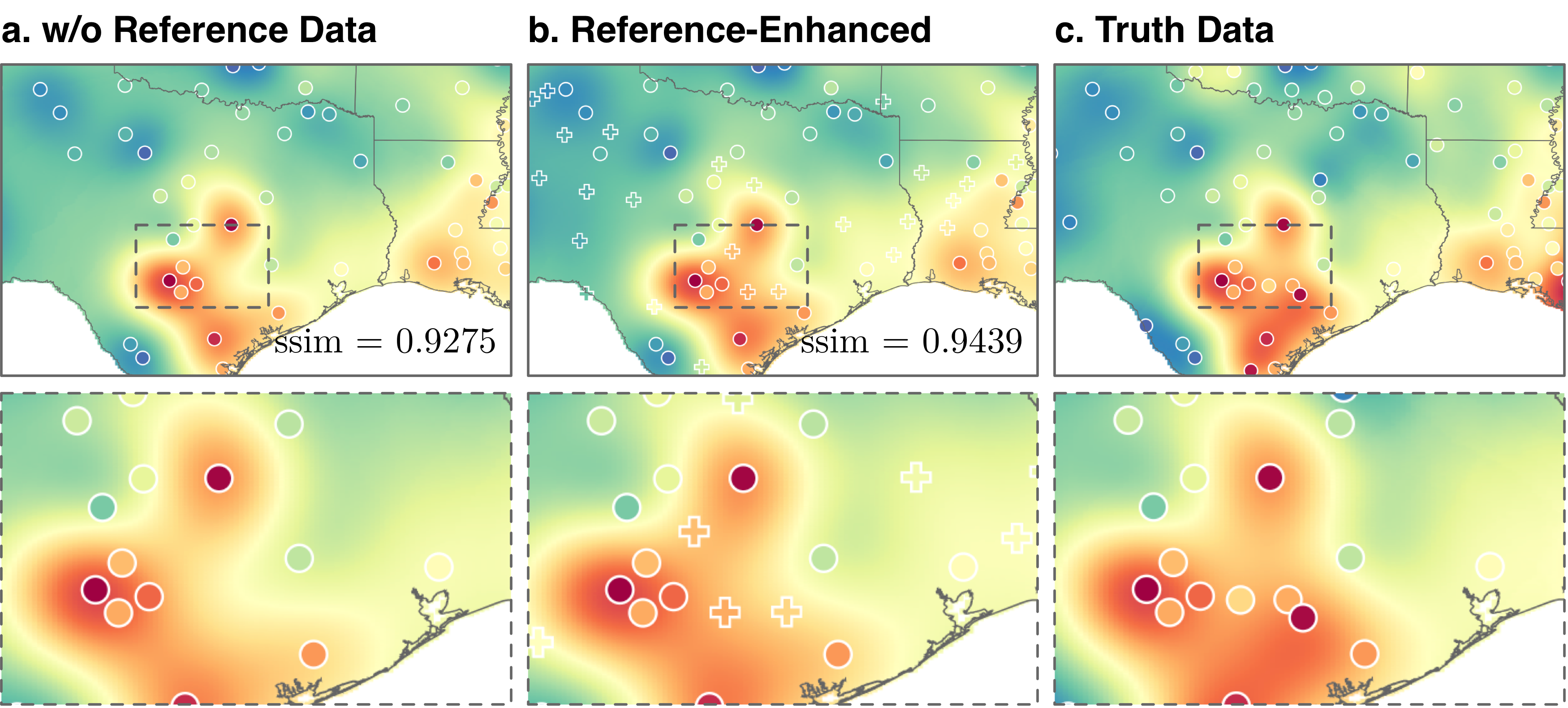}
  \caption{
    The interpolated heatmaps with (\bs{a}) or without (\bs{b}) imputation reference data, and with all sensor data (\bs{c}). Here we use the \texttt{ushcn} dataset, removing 40\% of sensors from the original sensor network. Incorporating imputation reference data leads to more accurate interpolation results, with increased SSIM compared to the truth data.
  }
  \label{fig:rbf}
\end{figure}

RBF interpolation solves $\*c$ and $\*b$ for all data points or a given maximum $N$ nearest neighbors to produce surface data. In our study, we use the Gaussian RBF function $\phi(r) = e^{(-\epsilon r^2)}$ for its smoothness and flexibility controlled by the shaping parameter $\epsilon$. We apply a small $\lambda = 0.5$ to avoid visual artifacts in the interpolant. We also constrained $N$ since larger values may cause quadratic growth in the amount of computational resources. The interpolant is subsequently color-mapped to obtain heatmaps.
\Cref{fig:rbf} compares the heatmaps interpolated on the original sensor network and using our reference-enhanced spatial interpolation method.

\section{Visualizing the Uncertainties}

Despite the availability of reference data, the heatmap contains various uncertainties. In this section, we identify three kinds of uncertainties and elaborate on our visualization design to communicate them.

\subsection{Sources of Uncertainty}
\label{sec:source}

Generating a heatmap involves sensor placement, data acquisition, and spatial interpolation. Each procedure introduces uncertainties and errors, potentially yielding inaccurate or misleading heatmaps \cite{prestonCommunicatingUncertaintyRisk2023,maceachrenVisualizingGeospatialInformation2005}. The primary sources of uncertainty include:

\begin{enumerate}[noitemsep,topsep=0pt,label={S\arabic*},leftmargin=*]
  \item \textbf{Sensor Placement}: Sensors are often unevenly distributed due to physical constraints or budgetary considerations. For instance, air quality sensors are usually concentrated in urban areas and sparsely installed in rural areas, and may not be accurately assessing values due to industrial emissions or wildfires, obscuring the true values.
  \item \textbf{Data Acquisition}: Prolonged deployment of sensors can expose them to various physical influences and compromise their performance, resulting in absent or abnormal values. While absent data can be easily identified, discerning between genuine anomalies and sensor-related issues is challenging. Abnormal values may be genuine unusual events, or they could be sensor or transmission errors.
  \item \textbf{Spatial Interpolation}: Spatial interpolation methods inherently are a process of estimation and approximation. These methods rely on available observations and parameter settings. Heatmaps generated under different parameters may exhibit substantial variances, which is a significant source of uncertainty requiring user awareness.
\end{enumerate}

\subsection{Visualization Design}

\begin{figure}[htbp]
  \centering
  \includegraphics[width=1\columnwidth]{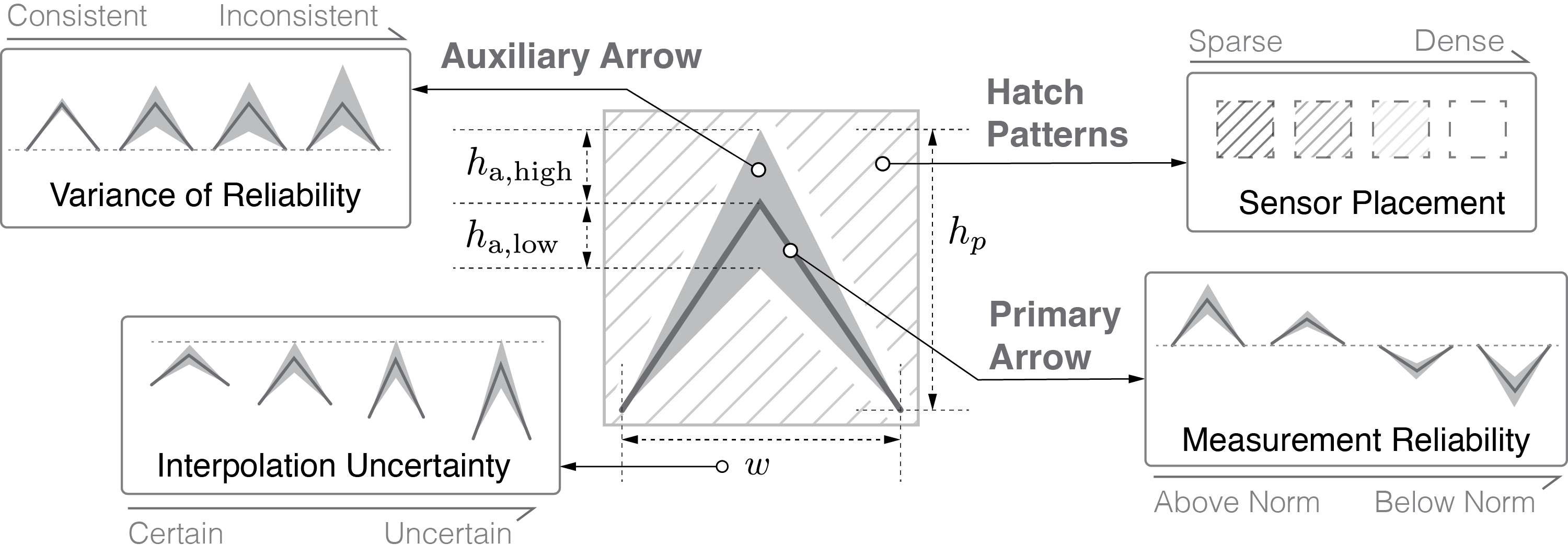}
  \caption{
    Visual encoding of RelMap. The hatch patterns in the background represent the sensor density distribution. Each reliability glyph has a width of $w$, containing a primary arrow (darker gray, line-shaped arrow in the center) with height $h_p$ and an auxiliary arrow (the lighter, arrowhead-shaped arrow in the background) with upper and lower heights $h_{\rm{high}, a}$ and $h_{\rm{low}, a}$. Various sources of uncertainties are encoded in the visualization. %
  }
  \label{fig:encoding}
\end{figure}

We design RelMap aiming at 1) effectively communicating the aforementioned sources of uncertainties, and 2) overlaying in conjunction with existing interpolation-based heatmaps. As shown in \cref{fig:encoding}, RelMap consists of two components:

\begin{enumerate}[noitemsep,topsep=0pt,label={\arabic*)},leftmargin=*]
  \item \textbf{Hatch Patterns}: A diagonal striped hatch texture pattern overlaying on the background heatmap, representing the spatial distribution of the original sensor network. 
  \item \textbf{Reliability Glyphs}: Arrow-like glyphs that convey the reliability of sensor values and the spatial interpolation dependencies. The original heatmap is evenly divided into $n$ grids $\{C_i\}_1^n$, with glyphs assigned on a per-grid basis to avoid visual clutter.
\end{enumerate}

\noindent
\textbf{Hatch Patterns}. We use opacity in hatch patterns to represent sensor placement, where regions with fewer sensors are more opaque, signaling higher uncertainty (\hyperref[sec:source]{S1}). To prevent the heatmap from becoming overloaded, we apply a threshold to the patterns, removing the patterns where sensor density is above a certain level. The sensor density distribution is derived through the KDE process detailed in \cref{sec:kde}.

\revision{\sidecomment{R1.1}
\noindent
\textbf{Reliability Glyphs}. We use the combination of primary arrows and auxiliary arrows, as shown in \cref{fig:encoding} to convey data acquisition reliability.
To quantify the uncertainty,
we first use our model to generate reference data. This is achieved by randomly masking half of the sensors to perform data imputation to get $X^*_A$, and repeating the process for the other half to get $X^*_B$. By merging $X^*_A$ and $X^*_B$, we obtain a reference value $X^*_{\rm{ref}}$ for each sensor, reflecting the model's interpretation of expected normal values. The reliability is quantified based on the deviation of observed values $X$ from $X^*_{\rm{ref}}$.} Then, the height of the primary arrow $h_p$ and the auxiliary arrows $h_a$ at grid $C_i$ are determined as follows:

\begin{equation}
  h_{p_i} = \frac{ \frac{1}{|C_j|} \sum_{j \in C_j} (X_j - X^*_{\rm{ref}_j})}{\max(h_a, h_p)}, h_{a_i} = \frac{Q(\{X_j - X_{\rm{ref}_j}^* | j \in C_j \})}{\max(h_a, h_p)},
\end{equation}

\noindent
where $Q$ denotes the lower quartile or upper quartile for the lower ($h_{\rm{low}, a}$) or upper ($h_{\rm{high}, a}$) auxiliary arrows, respectively (\hyperref[sec:source]{S3}). Inspired by box plots, the primary arrow height $h_p$ represents the average deviation of sensor readings from the expected values, while the auxiliary arrow height $h_a$ indicates the lower and upper quartiles of the deviations, indicating the potential existence of unreliable sensor readings within the cell area.

The width $w_i$ for each glyph encodes interpolation reliability. We use the average distance between sensors in the grid area and the grid center to determine the width of the glyph:

\begin{equation}
  w_i = 1 - \frac{\frac{1}{N}\sum_{j \in \Omega_i} \text{dist}(x - s_j)}{\max(w)}
\end{equation}

\noindent
where $N$ is the number of neighborhoods considered in the interpolation process. Areas with larger average distances exhibit narrower widths, producing sharper arrowheads to raise awareness of uncertainty introduced by spatial interpolation (\hyperref[sec:source]{S3}).

\begin{figure}[htbp]
  \centering
  \includegraphics[width=1\columnwidth]{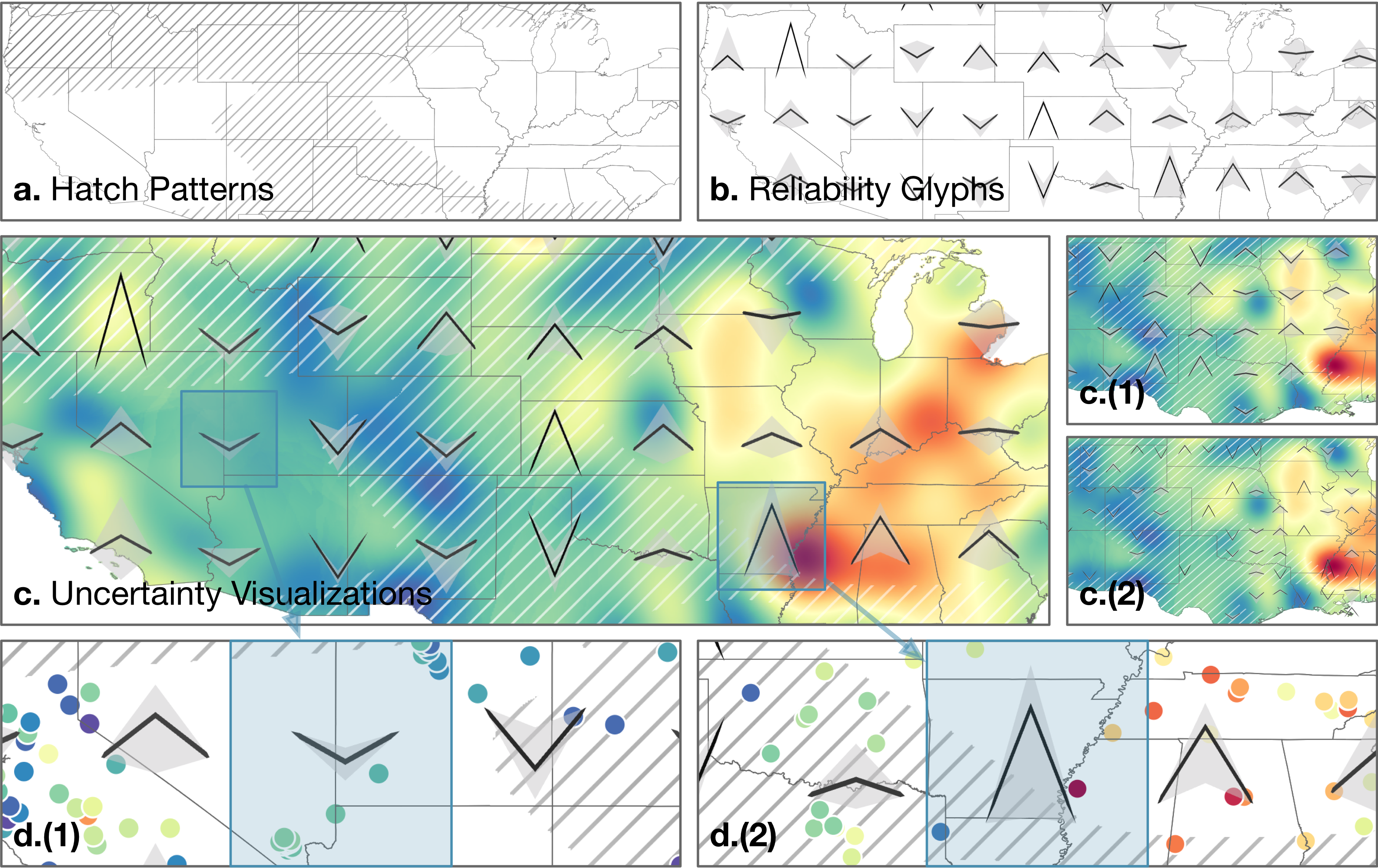}
  \caption{
    Example of RelMap visualization applied to the \texttt{uspm25} dataset, with Hatch Textures (\bs{a}) displaying sensor density and Reliability Glyphs (\bs{b}) indicating uncertainties within sensor observations and interpolation. The glyphs are placed on grids with configurable grid sizes (\bs{c.(1)} and \bs{c.(2)}). \bs{d.(1)} and \bs{d.(2)} shows the close-up view of two arrows. The sharper glyph head in \bs{d.(2)} signals larger distances of nearby sensors, indicating higher interpolation uncertainty. The auxiliary arrows in \bs{d.(2)} suggest the potential existence of unreliable sensor readings with abnormally high values.
  }
  \label{fig:reliamap}
\end{figure}

\sidecomment{R1.2}
\cref{fig:reliamap} displays a few examples of RelMap visualizations. \revision{%
We designate the hatch patterns and glyphs to use grayscale color encoding to avoid conflict with the heatmap's color scheme in the background. For areas with low sensor placement uncertainty, the texture fills are less visible, allowing a clear view of both the heatmap and the glyphs.
The glyph grid size is computed adaptively based on the users' zoom level and screen size to ensure readability.
}

\revision{
\subsection{Design Alternatives}
\label{sec:alt-design}
\sidecomment{R1.2\\R3.3}
During the design of RelMap, we have explored potential visual encodings and visualization techniques for presenting uncertainties.
Prior studies have proposed several methods for visualizing uncertainty in interpolation-based heatmaps, such as small multiples, ordered dotmaps, smoothed dotmaps, and risk contours \cite{kinkeldeyHowAssessVisual2014}. Small multiples allow multiple maps to be displayed simultaneously, but their effectiveness is limited on smaller screens. Variants of dotmaps divide each pixel into multiple cells and visualize multiple possible values sampled from the probabilistic distribution for that location. However, as intrinsic methods, they alter the original heatmap’s color encoding, which may hinder map interpretation \cite{prestonCommunicatingUncertaintyRisk2023}. Risk contours provide a more aggregated uncertainty overview but are unable to represent multiple sources of uncertainty simultaneously.
Therefore, we decide to use extrinsic glyphs to visualize uncertainties, facilitating ease of interpretation, deployment, and customization.

Regarding glyph placement, one alternative is to use Voronoi diagrams and place glyphs at the centroids of the cells. However, this may lead to visual clutter due to the irregular shapes of Voronoi regions. Additionally, it complicates zoom interactions, as glyph positions change unpredictably when new cells are generated. To ensure visual stability and consistency, we instead place glyphs on a regular grid, with cell sizes adaptively adjusted based on the current zoom level.
}

\section{Evaluation}

\newcommand{\evlfirst}[1]{\textbf{#1}}
\newcommand{\evlsecond}[1]{\setlength\fboxsep{1pt}\colorbox{LightGray}{#1}}

\label{sec:eval}

\revision{\sidecomment{R3.6}%
In this section, we evaluate the efficacy of our approach by validating each compoment of the framework: 1). data imputation accuracy, 2). spatial interpolation quality, and 3). temporal super-resolution quality, and 4). visual effectiveness. We also conduct an ablation study to assess the effectiveness of the PNA and GPE modules in our model. The evaluation is performed on four real-world geospatial sensor datasets, including
precipitation (\texttt{ushcn}, the U.S. Historical Climatology Network)\cite{menneUnitedStatesHistorical2012}, 1218 sensors),
PM2.5 (\texttt{uspm25}, from AirData \cite {unitedstatesenvironmentalprotectionagencyepaAirDataAir2016}, 749 sensors),
temperature (\texttt{catmp}, from AirData \cite {unitedstatesenvironmentalprotectionagencyepaAirDataAir2016}, 141 sensors),
and air quality index (\texttt{dtaqi}, Yangtze River Delta region, 294 sensors).
}

\revision{
The model is trained on each dataset separately, with hyperparameters available in the source code. To apply our method to a new dataset, re-training is required for the model to learn different spatiotemporal dependencies in the dataset.
\sidecomment{R1.3\\R3.5}%

}

\subsection{Data Imputation Accuracy}
\label{sec:eval-imputation}

We apply different mask rates $\alpha = \{0.2, 0.4, 0.5\}$ for the original data and apply different methods to generate imputation data. Following previous studies \cite{wuInductiveGraphNeural2020,applebyKrigingConvolutionalNetworks2020}, we use RMSE (Root Mean Square Error) and MAE (Mean Absolute Error) as metrics, computed only on the masked sensors. Three baseline methods are: 1). IGNNK \cite{wuInductiveGraphNeural2020}: a GNN-based approach using Diffusion Graph Convolution as the message-passing mechanism. 2). Ordinary Kriging (OKriging): a geo-statistical method that estimates values by fitting a variogram model, and 3). K-Nearest Neighbor (KNN): a deterministic method using the average of $k$ nearest neighbors as the target sensor value. In our experiment, we set $k = 5$.

\Cref{tab:imputation} shows the evaluation results. Our method's data imputation capability significantly outperforms IGNNK. This suggests that for spatiotemporal data imputation, our approach utilizing PNA convolution that collects messages directly in the spatial domain proves more advantageous than performing aggregation in the spectrum domain. Compared with traditional methods, our method performs better in most cases, except for \texttt{dtaqi} and \texttt{ushcn} at the mask rate of $\alpha = 0.2$. Here, OKriging achieves better performance, indicating that when the task load for imputation is light, traditional methods remain capable of modeling spatial correlations. However, as $\alpha$ increases, OKriging becomes less effective. This stems from the incorporation of both spatial and temporal information in our approach, enabling it to learn more complex semantic information.

\begin{table*}[tbp]
     \caption{
         The performance of different methods on the data imputation task. Here $\alpha$ denotes the mask rate. The lower and upper range displayed here is the 0.1 and 99.9 percentile of the data. The best results are highlighted with \evlfirst{bold} fonts and the second best with \evlsecond{gray} background.
     }
     \label{tab:imputation}
     \renewcommand\arraystretch{0.68}
     \centering
     \small

\begin{tabular}{ccccccccccccccc}
     \toprule
     \multirow{2}[2]{*}{Dataset} & \multirow{2}{*}{\shortstack{Range \\ Unit}} & \multirow{2}[2]{*}{$\alpha$} & \multicolumn{2}{c}{\textbf{ImTerp}} & \multicolumn{2}{c}{\textbf{IGNNK}\cite{wuInductiveGraphNeural2020}} & \multicolumn{2}{c}{\textbf{OKriging}} & \multicolumn{2}{c}{\textbf{KNN}} & \multicolumn{2}{c}{\textbf{ImTerp w/o GPE}} & \multicolumn{2}{c}{\textbf{ImTerp w/o PNA}} \\
          &      &      & RMSE $\downarrow$ & MAE $\downarrow$ & RMSE $\downarrow$ & MAE $\downarrow$ & RMSE $\downarrow$ & MAE $\downarrow$ & RMSE $\downarrow$ & MAE $\downarrow$ & RMSE $\downarrow$ & MAE $\downarrow$ & RMSE $\downarrow$ & MAE $\downarrow$ \\
     \midrule
     \multirow{3}{*}{\texttt{ushcn}} & \multicolumn{1}{c}{\multirow{3}{*}{\shortstack{[0.0, 67.2] \\ mm }}} & 0.2  & \evlsecond{1.878} & \evlsecond{0.662} & 3.280 & 1.163 & \evlfirst{1.772} & \evlfirst{0.495} & 5.562 & 1.903 & 2.352 & 0.891 & 2.484 & 1.005 \\
          &      & 0.4  & \evlfirst{2.233} & \evlfirst{0.764} & 3.115 & 1.269 & 2.507 & 1.004 & 7.584 & 3.707 & \evlsecond{2.413} & \evlsecond{0.939} & 2.416 & 0.996 \\
          &      & 0.5  & \evlfirst{2.083} & \evlfirst{0.768} & 3.681 & 1.590 & 2.813 & 1.269 & 8.311 & 4.548 & \evlsecond{2.341} & \evlsecond{0.923} & 2.509 & 1.088 \\
     \midrule
     \multirow{3}{*}{\texttt{uspm25}} & \multicolumn{1}{c}{\multirow{3}{*}{\shortstack{[0.0, 214.5] \\ \textmu g/m$^{3}$ }}} & 0.2  & \evlfirst{4.035} & 1.263 & 4.767 & 1.556 & 4.708 & \evlfirst{0.844} & 5.698 & \evlsecond{1.237} & \evlsecond{4.647} & 1.730 & 5.007 & 1.925 \\
          &      & 0.4  & \evlfirst{4.040} & \evlfirst{1.274} & 4.362 & \evlsecond{1.544} & 5.853 & 1.712 & 7.520 & 2.494 & 4.355 & 1.625 & \evlsecond{4.342} & 1.645 \\
          &      & 0.5  & \evlfirst{4.031} & \evlfirst{1.273} & 4.912 & 1.995 & 6.186 & 2.117 & 8.015 & 3.074 & 4.577 & 1.718 & \evlsecond{4.441} & \evlsecond{1.700} \\
     \midrule
     \multirow{3}{*}{\texttt{catmp}} & \multicolumn{1}{c}{\multirow{3}{*}{ \shortstack{[-35.0, 132.1] \\ $^{\circ}$F }}} & 0.2  & \evlfirst{7.170} & \evlfirst{1.487} & 13.122 & 6.568 & 11.939 & 3.557 & 12.878 & 3.891 & 7.661 & \evlsecond{1.508} & \evlsecond{7.627} & 1.638 \\
          &      & 0.4  & \evlfirst{7.765} & \evlfirst{1.604} & 12.862 & 4.401 & 15.502 & 6.066 & 16.839 & 7.187 & 8.152 & \evlsecond{1.613} & \evlsecond{7.899} & 1.672 \\
          &      & 0.5  & \evlfirst{5.871} & \evlfirst{1.269} & 14.531 & 5.273 & 16.847 & 7.151 & 18.421 & 8.719 & 8.045 & \evlsecond{1.549} & \evlsecond{7.579} & 1.647 \\
     \midrule
     \multirow{3}{*}{\texttt{dtaqi}} & \multicolumn{1}{c}{\multirow{3}{*}{ \shortstack{[0, 462.7] \\ IAQI }}} & 0.2  & \evlsecond{9.829} & \evlsecond{3.408} & 13.309 & 5.269 & \evlfirst{7.799} & \evlfirst{2.206} & 14.645 & 5.099 & 14.070 & 5.693 & 11.839 & 5.070 \\
          &      & 0.4  & \evlfirst{10.568} & \evlfirst{3.782} & 11.933 & 4.865 & 11.527 & 4.720 & 20.289 & 9.96 & 12.050 & 4.707 & \evlsecond{10.666} & \evlsecond{4.482} \\
          &      & 0.5  & \evlsecond{12.542} & \evlsecond{4.457} & 16.037 & 6.790 & 12.990 & 5.960 & 22.622 & 12.396 & 10.964 & \evlfirst{4.293} & \evlfirst{12.008} & 5.107 \\
     \bottomrule
     \end{tabular}%
 \end{table*}

\subsection{Spatial Interpolation Quality}
\label{sec:eval-tsr}

\revision{%
\sidecomment{R2.4\\R3.6}%
To validate the effectiveness of the imputation reference data in enhancing spatial interpolation quality, we use a fixed masked rate of $\alpha = 0.3$ on the full data and use different densification ratios $\delta = \{0, 0.2, 0.4\}$ to impute reference data on densified sensors. Then, we interpolate four heatmaps using the same KDE parameters ($\epsilon = 1.0$, $n = 10$): the reference heatmap using full data, and the maps generated under different $\delta$ values with reference data incorporated. Then, we compare the SSIM (Structural Similarity) score of the interpolated heatmaps with the truth map.
}

\begin{figure}[htbp]
     \centering
     \includegraphics[width=1\columnwidth]{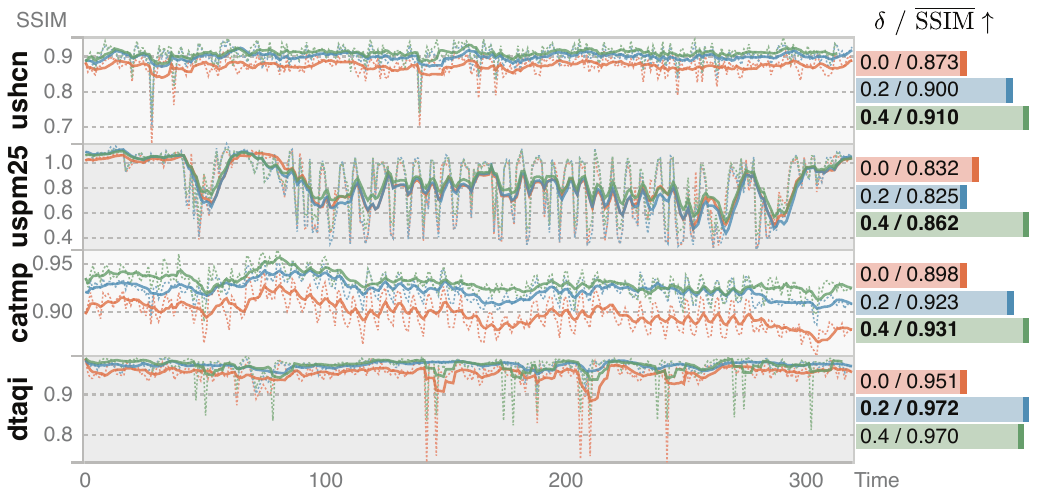}
     \caption{
        Spatial interpolation quality comparison with different densification ratios ($\delta$). We test the first 320 timesteps for each dataset, displaying the SSIM of the interpolated heatmap with the truth heatmap. To make the trends more readable, we apply a smoothed moving average with a window size of 5. The actual values are shown with dashed lines.
     }
     \label{fig:eval-interp}
\end{figure}

\Cref{fig:eval-interp} illustrates the result. For all datasets, the incorporation of densified sensors improved the interpolation results. In some cases where the masked sensors contain critical information, the heatmap without imputation reference will undergo a drastic drop in SSIM, indicating a significant deviation from the truth, such as time 146 for the \texttt{ushcn} dataset, as shown in \cref{fig:rbf}. For dataset \texttt{uspm25}, $\delta = 0.2$ causes the heatmap to be slightly less accurate, which may be due to the densification process placing virtual sensors at the periphery of the sample area. When further densifying the sensors, the interpolation quality improves, as sensors are located in more informative areas. In general, the interpolant can better capture the spatial characteristics in the areas of interest with reference data and produce heatmaps that better reflect actual situations.

\subsection{Temporal Super-Resolution Quality}

We evaluate the temporal SR quality with different SR rates $\rm{SR} = \times \{2, 4, 8\}$. We temporally downscale the data by SR times as input and use a fixed mask rate of $\alpha = 0.3$. We let the model impute the masked data and perform temporal SR simultaneously, and compute the RMSE and MAE with the full data. Our approach is compared with a baseline method that generates missing data using piece-wise linear interpolation.

\begin{table}[htbp]
    \small
    \caption{
        Temporal super-resolution quality under different SR rates. The best result for each rate is highlighted with \evlfirst{bold} font and the second best result with \evlsecond{gray} background.
    }
    \renewcommand\arraystretch{0.68}
    \setlength{\tabcolsep}{2pt}
    \centering
    \label{tab:tsr}

\begin{tabular}{cccccccccc}
     \toprule
     \multirow{2}[2]{*}{Dataset} & \multirow{2}[2]{*}{SR} & \multicolumn{2}{c}{\textbf{ImTerp}} & \multicolumn{2}{c}{\textbf{Linear}} & \multicolumn{2}{c}{\textbf{ImTerp w/o GPE}} & \multicolumn{2}{c}{\textbf{ImTerp w/o PNA}} \\
          &      & RMSE $\downarrow$ & MAE $\downarrow$ & RMSE $\downarrow$ & MAE $\downarrow$ & RMSE $\downarrow$ & MAE $\downarrow$ & RMSE $\downarrow$ & MAE $\downarrow$ \\
     \midrule
     \multirow{3}{*}{\texttt{ushcn}} & $\times 2$ & \evlfirst{2.996} & \evlfirst{1.119} & 5.619 & 3.320 & \evlsecond{3.103} & 1.275 & 3.114 & \evlsecond{1.249} \\
          & $\times 4$ & \evlfirst{3.339} & \evlfirst{1.303} & 6.522 & 4.197 & \evlsecond{3.528} & \evlsecond{1.520} & 3.992 & 1.657 \\
          & $\times 8$ & \evlfirst{3.325} & \evlfirst{1.323} & 6.902 & 4.638 & \evlsecond{3.792} & 1.633 & 3.783 & \evlsecond{1.587} \\
     \midrule
     \multirow{3}{*}{\texttt{uspm25}} & $\times 2$ & \evlfirst{4.114} & \evlfirst{1.224} & \evlsecond{5.597} & 2.324 & 6.601 & \evlsecond{1.375} & 7.907 & 1.974 \\
          & $\times 4$ & \evlfirst{4.449} & \evlfirst{1.328} & 6.514 & 3.074 & 8.031 & 1.529 & 8.846 & 1.872 \\
          & $\times 8$ & \evlfirst{4.272} & \evlfirst{1.271} & 7.878 & 3.807 & \evlsecond{6.233} & \evlsecond{1.442} & 6.948 & 1.553 \\
     \midrule
     \multirow{3}{*}{\texttt{catmp}} & $\times 2$ & \evlfirst{6.598} & \evlfirst{1.294} & 8.546 & 4.848 & \evlsecond{7.536} & 1.423 & 7.892 & \evlsecond{1.404} \\
          & $\times 4$ & \evlsecond{7.402} & \evlfirst{1.317} & 10.241 & 6.562 & \evlfirst{6.928} & \evlsecond{1.437} & 8.438 & 1.665 \\
          & $\times 8$ & \evlsecond{7.867} & 1.587 & 11.018 & 7.257 & \evlfirst{6.830} & \evlfirst{1.417} & 8.439 & \evlsecond{1.577} \\
     \midrule
     \multirow{3}{*}{\texttt{dtaqi}} & $\times 2$ & \evlfirst{11.281} & \evlfirst{3.997} & \evlsecond{12.674} & 6.205 & 12.809 & 4.481 & 12.910 & \evlsecond{4.345} \\
          & $\times 4$ & \evlfirst{11.186} & \evlfirst{3.978} & 17.030 & 9.773 & 13.407 & 4.706 & \evlsecond{13.375} & \evlsecond{4.326} \\
          & $\times 8$ & \evlfirst{11.954} & \evlfirst{4.668} & 19.626 & 12.728 & \evlsecond{12.816} & 5.186 & 14.222 & \evlsecond{5.181} \\
     \bottomrule
     \end{tabular}%

\end{table}

The results are listed in \cref{tab:tsr}. We can observe that our approach has improvements ranging from 10\% to 90\%, indicating our method can better characterize the non-linear transitions between time steps. As the SR rate increases, the amount of reference information decreases, leading to a declining trend for the TSR quality, except for \texttt{uspm25} at the $\times 8$ resolution, which has a lower RMSE compared with $\times 4$. This may be attributed to the data exhibiting periodic changes, leading to better results under this SR rate. The results suggest that our approach can better capture the non-linear transitions during TSR.

\begin{figure}[thbp]
     \centering
     \includegraphics[width=0.95\columnwidth]{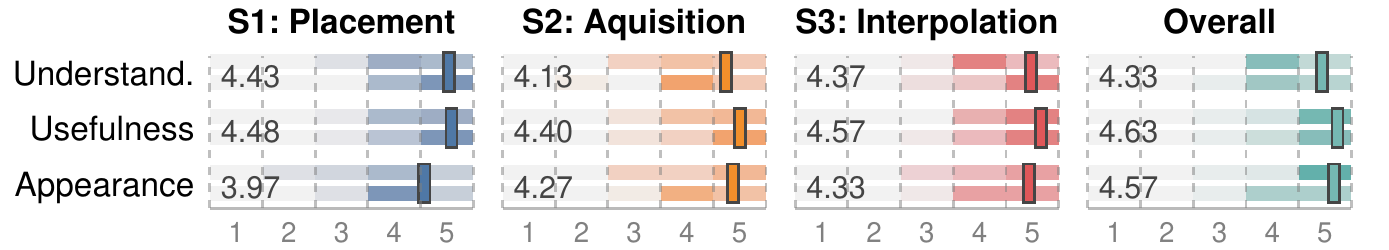}
     \caption{
     Compiled results of the user study questionnaire. The displayed score is the average of the two questions assessing the corresponding evaluation perspective and uncertainty source.
     }
     \label{fig:eval-user}
\end{figure}

\subsection{Ablation Study}

To assess the effectiveness of PNA and GPE modules in our model, we conduct an ablation study by removing the two components separately and testing the imputation and TSR performance. When PNA is not applied, we replace it with a single \ttt{MEAN} aggregator and an identity scaler. The results are listed in the corresponding columns in \cref{tab:imputation} and \cref{tab:tsr}.

The PNA module significantly enhances the model's ability to capture spatial dependencies during convolution, benefiting most datasets and tasks except for the TSR task on \texttt{catmp}. This dataset has fewer sensors (141), suggesting PNA might suffer from small sensor counts, leading to overfitting. GPE, as an approach to input augmentation,
generally improving performance across various scenarios. Our results show that GPE is as important as PNA, given that a model without either PNA or GPE produces similar second-best performances. However, for the imputation task of the \texttt{catmp} dataset, the performance improves without GPE. This is due to the dataset's highly uneven sensor distribution, with 80\% of sensors concentrated in 10\% of the sampling domain, posing challenges for GPE in learning geographic semantics across locations.

\begin{figure*}[bthp]
     \centering
     \includegraphics[width=0.88\textwidth]{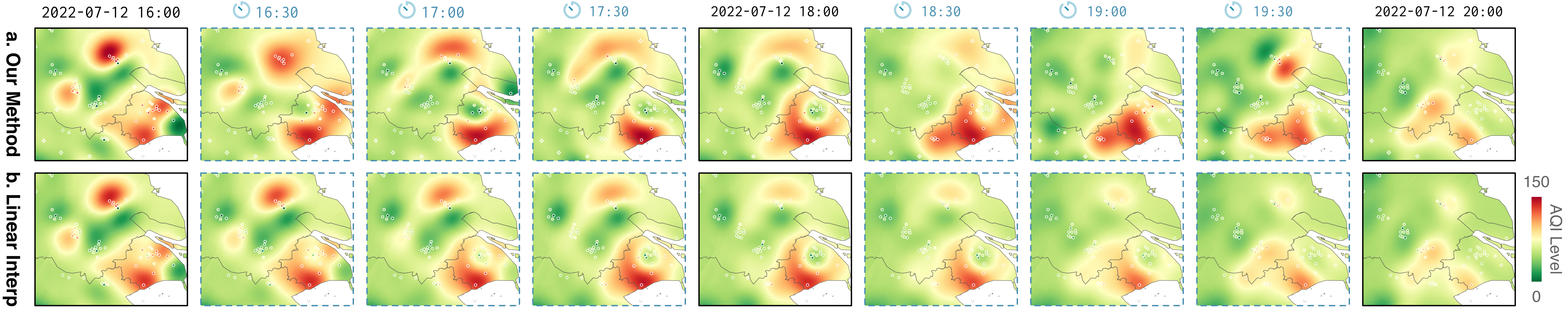}
     \caption{Temporal super-resolution applied to the \texttt{dtaqi} data. The temporal resolution is increased from 2 hours to 30 minutes. The heatmaps produced by super-resolution are displayed with dashed blue borders. Our approach reveals a more informative dynamic trend compared with linear interpolation.
     }
     \label{fig:tsr}
\end{figure*}

\subsection{Visual Effectiveness}

We conduct a user study to evaluate our visualization from three perspectives: 1). Understandability: if the user can comprehend how the visual encodings are mapped to different uncertainties; 2). Usefulness: if the user finds the visualizations helpful in communicating uncertainties and map interpretation; 3). Appearance: whether the visualizations are well integrated with the heatmaps and visually appealing. 15 participants (age $\mu = 23.4, \sigma = 2.85$), including 45\% not familiar with spatiotemporal data visualization, were recruited from the campus. We first introduced our design and provided them with a set of heatmaps with our uncertainty visualizations. Participants were asked to rate the visualizations on three sections of 5-point Likert scales, each corresponding to one source of uncertainty. Each section included six questions, assessing one of the three perspectives. We also collected verbal feedback in the process.

\Cref{fig:eval-user} shows the questionnaire results. Most participants agree that the visualizations are easy to understand (4.33 / 5 overall understandability) and find them useful in interpreting the heatmaps, particularly in conveying uncertainty in spatial interpolation (4.57 / 5 usefulness) and sensor placement (4.48 / 5 usefulness). One participant motioned that the sharper glyphs could immediately raise awareness of uncertainty and recognize that the data in this area may not be reliable. Participants also agree that our visualization does not interfere with map interpretation, and expressed their interest in applying these visualizations in everyday applications, like weather forecasts (11 / 15 agree). However, some participants (2 / 15) pointed out the potential for visual clutter when textures overlaid on the road networks. This can be addressed by adjusting the density and transparency of the textures to reduce visual interference.

\section{Application Scenarios}

\subsection{Reliable Spatial Interpolation}

\begin{figure}[hbt]
  \centering
  \includegraphics[width=1\columnwidth]{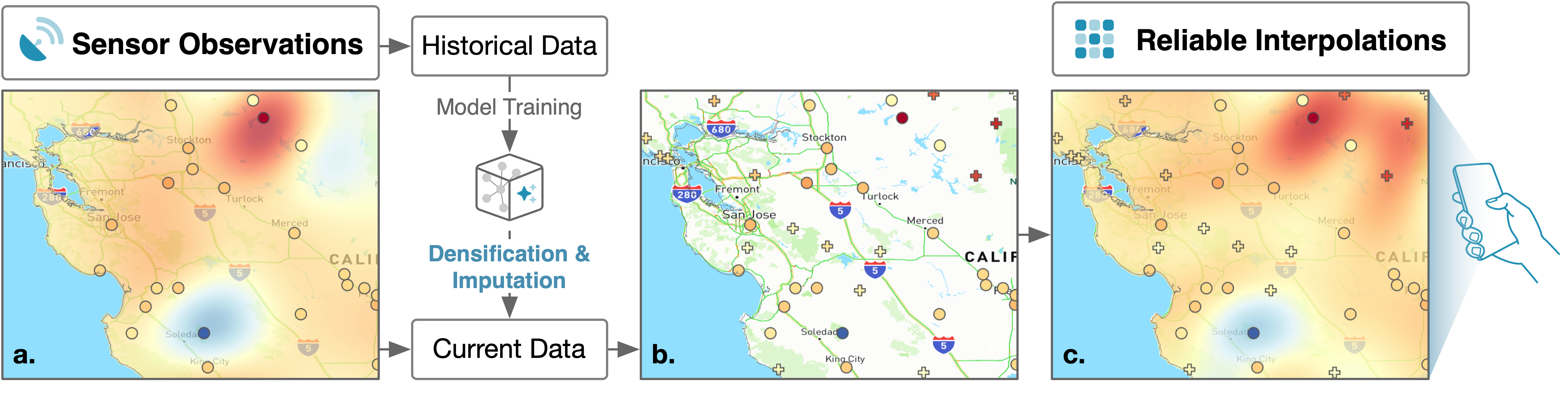}
  \caption{
    Example of generating more reliable heatmaps. Existing observations (\bs{a}) will be used for training to learn the spatiotemporal dependencies. Our model performs data imputation (\bs{b}) on densified sensors to generate more reliable heatmaps for user interpretation (\bs{c}).
  }
  \label{fig:app-reliable}
\end{figure}

\begin{cjte}
Applying our approach to the \hyperref[sec:eval]{\texttt{catmp}} dataset, we demonstrate our approach's capability to generate more reliable heatmaps for sensors, especially in areas with sparse coverage. As shown in \cref{fig:app-reliable}, a region with sparse sensor coverage exists in the upper right corner of the heatmap, located in Central California, west of Yosemite. Directly applying spatial interpolation leads to a heatmap with a lower temperature in this area. With our approach, virtual sensors are evenly distributed in this region, and their data are imputed using the latest sensor readings. The resulting heatmap reveals the elevated temperature in this area, as guided by the imputation references marked with plus symbols (\cref{fig:app-reliable}.c). This could be the result of the model learning historical information and recognizing similar spatiotemporal patterns in this area. Conversely, in regions already covered by sensors, such as the lower area of the map, fewer virtual sensors are added, and the resulting heatmap remains similar.

\end{cjte}

\subsection{Informative Temporal Super-resolution for Heatmaps}

\begin{cjte}
  In this application, we demonstrate the temporal super-resolution (TSR) capability using the \hyperref[sec:eval]{\texttt{dtaqi}} dataset. As shown in \cref{fig:tsr}, by applying $\times 4$ super-resolution, the temporal resolution is increased from 2 hours to 30 minutes, allowing users to obtain air quality data with more detailed time steps and facilitate activity planning. \cref{fig:tsr}.a and \cref{fig:tsr}.b show the TSR results of our approach and linear interpolation, respectively. Heatmaps produced by our approach reveal a southwest movement of polluted areas between 16:00 to 20:00, while linear interpolation only reveals a gradual change. Our results also highlight a raised region at 19:30 in the upper middle section, which is not detectable through linear interpolation. Our model is supervised to generate informative, rather than smooth transitions during masked subgraph training, allowing it to fill in the missing data with more informative dynamic patterns.
\end{cjte}

\subsection{Uncertainty-aware Visualizations}

\begin{cjte}
To demonstrate the effectiveness of our uncertainty visualization, we apply our design to \hyperref[sec:eval]{\texttt{uspm25}} and \hyperref[sec:eval]{\texttt{dtaqi}} dataset, where \ttt{dtaqi} dataset exhibits more uneven sensor distributions. For \ttt{uspm25}, in the southern Texas region, the heatmap displays elevated levels of PM2.5, with the height of the primary arrow being low and areas spanned by the auxiliary arrows being small. This suggests that the interpolated data here are highly certain and reliable, supported by a large number of sensors offering consistent observations. The recent wildfires near the Houston area might have contributed to the increased pollution levels. Conversely, for the \ttt{dtaqi} dataset, the area around North Zhejiang Province also shows elevated values. However, the narrower width, sharper glyph head, and larger area spanned by auxiliary arrows indicate that sensor observations in this region are inconsistent, hinting that the actual AQI levels may not be as critical as the heatmap suggests. Additionally, the presence of hatch patterns in the background signals sparse sensor placement, reminding users of the uncertainty in this unreliable interpolation. Though both heatmaps display elevated values, RelMap offers more comprehensive understanding of the data.
\end{cjte}

\begin{figure}[htbp]
  \centering
  \includegraphics[width=1\columnwidth]{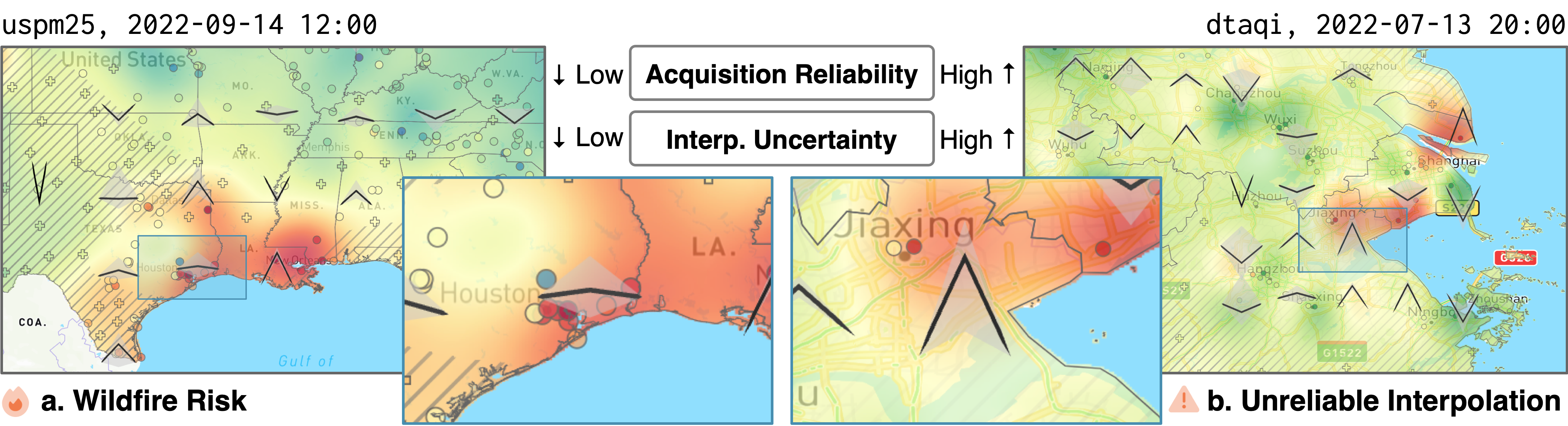}
  \caption{
    Examples for uncertainty-aware visualizations. The glyphs in \bs{a} have a smooth head and small auxiliary arrow regions, indicating a low level of uncertainty for data acquisition and interpolation. In contrast to \bs{b}, the sharp head and large auxiliary arrow cover range indicate high uncertainties with sparse sensor placement.
  }
  \label{fig:app-vis}
\end{figure}


\vspace{-12pt}
\section{Discussion and Conclusion}

\sidecomment{R2.1\\R3.4}
\revision{
\noindent
\textbf{Scalability.}
On the \textit{computational side}, we model the spatial relationship using a dense graph with spatial complexity of $O(n^2)$. Every PNA aggregation layer touches only $O(kn)$ edges, keeping both training and inference linear in the number of sensors. Combined with masked subgraph sampling and temporal windowing, training with thousands of sensors can be completed within hours on a single RTX 4070 GPU. However, for larger sensor networks, computation bottleneck may occur and requires techniques such as graph scarification and distributed training.
On the \textit{visualization side}, we employ density-aware hatch patterns and grid-based glyph design. As the resolution or spatial coverage increases, coarsening the grid and adjusting density thresholds can be used to maintain the clarity. In practical deployment, different alpha can be applied to different layers, making RelMap scalable to city-to-national scale sensor networks.
}

\revision{
\noindent
\textbf{Limitations.}\sidecomment{R2.2}
Though promising, our approach has some limitations. First, it currently only supports scalar data, while vector data such as wind fields and ocean currents also play a crucial role in many applications. Extending our model to vector data is a future direction of research. Second, our model only considers univariate information for imputation. Leveraging multivariate data as an additional modality could lead to more accurate interpolation results. For instance, combining remote sensing images with sensor sampling may produce more accurate heatmaps for air quality. Lastly, the uncertainty information quantified by our approach is on a per-sensor basis, which may not fully capture the actual scenarios. Future work could explore more sophisticated uncertainty modeling techniques that account for spatial dependencies.

In this paper, we propose RelMap, a novel pipeline for visualizing spatiotemporal sensor data that first augment the reliability of interpolation using imputation reference data produced by a GNN-based model, and then visualize various uncertainties on the heatmap. Our method makes heatmaps uncertainty-aware, supporting users in better interpreting the data and making informed decisions. The source code is available at
\url{https://github.com/jtchen2k/relmap}.
}


\acknowledgments{
The authors wish to acknowledge the support from the Natural Science Foundation of Shanghai Municipality, China under Grant 24ZR1418300.
}

\bibliographystyle{abbrv-doi-hyperref-narrow}

\bibliography{ref_final}

\begin{thebibliography}{10}
\renewcommand*{\sfdefault}{PTSansNarrow-TLF}

\bibitem{applebyKrigingConvolutionalNetworks2020}
G.~Appleby, L.~Liu, and L.-P. Liu.
\newblock {Kriging {{Convolutional Networks}}}.
\newblock {\em Proceedings of the AAAI Conference on Artificial Intelligence},
  34(04):3187--3194, 2020. \href{https://doi.org/10.1609/aaai.v34i04.5716}
{doi: \textsf{%
10\hspace{.1pt}\discretionary{.}{%
}{.}\hspace{.4pt}1609\discretionary{/}{%
}{/}aaai\hspace{.1pt}\discretionary{.}{%
}{.}\hspace{.4pt}v34i04\hspace{.1pt}\discretionary{.}{%
}{.}\hspace{.4pt}5716}}


\bibitem{atwoodDiffusionConvolutionalNeuralNetworks}
J.~Atwood and D.~Towsley.
\newblock {Diffusion-Convolutional Neural Networks}.
\newblock In {\em {Proceedings of the 30th International Conference on Neural
  Information Processing Systems (NIPS' 16)}}, pp. 1993--2001, 2016.

\bibitem{brewerDesigningBetterMaps2015}
C.~Brewer.
\newblock {\em {Designing {{Better Maps}}: {{A Guide}} for {{GIS Users}}}}.
\newblock 2nd ed., 2015.

\bibitem{brunaSpectralNetworksLocally2014}
J.~Bruna, W.~Zaremba, A.~Szlam, and Y.~LeCun.
\newblock {Spectral {{Networks}} and {{Locally Connected Networks}} on
  {{Graphs}}}.
\newblock In {\em {2nd International Conference on Learning Representations,
  {ICLR} 2014}}, 2014.

\bibitem{caoVoilaVisualAnomaly2018}
N.~Cao, C.~Lin, Q.~Zhu, Y.-R. Lin, X.~Teng, and X.~Wen.
\newblock {Voila: {{Visual Anomaly Detection}} and {{Monitoring}} with
  {{Streaming Spatiotemporal Data}}}.
\newblock {\em IEEE Transactions on Visualization and Computer Graphics},
  24(1):23--33, 2018. \href{https://doi.org/10.1109/TVCG.2017.2744419}
{doi: \textsf{%
10\hspace{.1pt}\discretionary{.}{%
}{.}\hspace{.4pt}1109\discretionary{/}{%
}{/}TVCG\hspace{.1pt}\discretionary{.}{%
}{.}\hspace{.4pt}2017\hspace{.1pt}\discretionary{.}{%
}{.}\hspace{.4pt}2744419}}


\bibitem{chenVFDPVisualAnalysis2022}
C.~Chen, C.~Li, J.~Chen, and C.~Wang.
\newblock {{{VFDP}}: {{Visual Analysis}} of {{Flight Delay}} and
  {{Propagation}} on a {{Geographical Map}}}.
\newblock {\em IEEE Transactions on Intelligent Transportation Systems},
  23(4):3510--3521, 2022. \href{https://doi.org/10.1109/TITS.2020.3037191}
{doi: \textsf{%
10\hspace{.1pt}\discretionary{.}{%
}{.}\hspace{.4pt}1109\discretionary{/}{%
}{/}TITS\hspace{.1pt}\discretionary{.}{%
}{.}\hspace{.4pt}2020\hspace{.1pt}\discretionary{.}{%
}{.}\hspace{.4pt}3037191}}


\bibitem{chen2022adaptiveAGCIN}
F.~Chen, D.~Wang, S.~Lei, J.~He, Y.~Fu, and C.-T. Lu.
\newblock {Adaptive graph convolutional imputation network for environmental
  sensor data recovery}.
\newblock 10:1025268. \href{https://doi.org/10.3389/fenvs.2022.1025268}
{doi: \textsf{%
10\hspace{.1pt}\discretionary{.}{%
}{.}\hspace{.4pt}3389\discretionary{/}{%
}{/}fenvs\hspace{.1pt}\discretionary{.}{%
}{.}\hspace{.4pt}2022\hspace{.1pt}\discretionary{.}{%
}{.}\hspace{.4pt}1025268}}


\bibitem{chen2024salientime}
J.~Chen, H.~Huang, H.~Ye, P.~Zhong, C.~Li, and C.~Wang.
\newblock {{{SalienTime}}: {{User-driven}} Selection of Salient Time Steps for
  Large-Scale Geospatial Data Visualization}.
\newblock In {\em {Proceedings of the 2024 {{CHI}} Conference on Human Factors
  in Computing Systems}}, 2024. \href{https://doi.org/10.1145/3613904.3642944}
{doi: \textsf{%
10\hspace{.1pt}\discretionary{.}{%
}{.}\hspace{.4pt}1145\discretionary{/}{%
}{/}3613904\hspace{.1pt}\discretionary{.}{%
}{.}\hspace{.4pt}3642944}}


\bibitem{chenSenseMapUrbanPerformance2023}
J.~Chen, Q.~Huang, C.~Wang, and C.~Li.
\newblock {SenseMap: Urban Performance Visualization and Analytics Via Semantic
  Textual Similarity}.
\newblock {\em IEEE Transactions on Visualization and Computer Graphics},
  30(9):6275--6290, 2024. \href{https://doi.org/10.1109/TVCG.2023.3333356}
{doi: \textsf{%
10\hspace{.1pt}\discretionary{.}{%
}{.}\hspace{.4pt}1109\discretionary{/}{%
}{/}TVCG\hspace{.1pt}\discretionary{.}{%
}{.}\hspace{.4pt}2023\hspace{.1pt}\discretionary{.}{%
}{.}\hspace{.4pt}3333356}}


\bibitem{cheongEvaluatingImpactVisualization2016}
L.~Cheong, S.~Bleisch, A.~Kealy, K.~Tolhurst, T.~Wilkening, and M.~Duckham.
\newblock {Evaluating the Impact of Visualization of Wildfire Hazard upon
  Decision-Making under Uncertainty}.
\newblock {\em International Journal of Geographical Information Science},
  30(7):1377--1404, 2016. \href{https://doi.org/10.1080/13658816.2015.1131829}
{doi: \textsf{%
10\hspace{.1pt}\discretionary{.}{%
}{.}\hspace{.4pt}1080\discretionary{/}{%
}{/}13658816\hspace{.1pt}\discretionary{.}{%
}{.}\hspace{.4pt}2015\hspace{.1pt}\discretionary{.}{%
}{.}\hspace{.4pt}1131829}}


\bibitem{cini2022fillingGRIN}
A.~Cini, I.~Marisca, and C.~Alippi.
\newblock {Filling the G\_ap\_s: Multivariate Time Series Imputation by Graph
  Neural Networks}. \href{https://doi.org/10.48550/arXiv.2108.00298}
{doi: \textsf{%
10\hspace{.1pt}\discretionary{.}{%
}{.}\hspace{.4pt}48550\discretionary{/}{%
}{/}arXiv\hspace{.1pt}\discretionary{.}{%
}{.}\hspace{.4pt}2108\hspace{.1pt}\discretionary{.}{%
}{.}\hspace{.4pt}00298}}


\bibitem{corsopna}
G.~Corso, L.~Cavalleri, D.~Beaini, P.~Liò, and P.~Veličković.
\newblock {{Principal Neighbourhood Aggregation for Graph Nets}}.
\newblock In {\em {Proceedings of the 34th International Conference on Neural
  Information Processing Systems (NIPS' 20)}}.
  \href{https://doi.org/10.5555/3495724.3496836}
{doi: \textsf{%
10\hspace{.1pt}\discretionary{.}{%
}{.}\hspace{.4pt}5555\discretionary{/}{%
}{/}3495724\hspace{.1pt}\discretionary{.}{%
}{.}\hspace{.4pt}3496836}}


\bibitem{dauphinLanguageModelingGated2017}
Y.~N. Dauphin, A.~Fan, M.~Auli, and D.~Grangier.
\newblock {Language Modeling with Gated Convolutional Networks}.
\newblock In {\em {Proceedings of the 34th International Conference on Machine
  Learning}}, vol.~70, pp. 933--941, 2017.
  \href{https://doi.org/10.5555/3305381.3305478}
{doi: \textsf{%
10\hspace{.1pt}\discretionary{.}{%
}{.}\hspace{.4pt}5555\discretionary{/}{%
}{/}3305381\hspace{.1pt}\discretionary{.}{%
}{.}\hspace{.4pt}3305478}}


\bibitem{defferrardConvolutionalNeuralNetworks2016}
M.~Defferrard, X.~Bresson, and P.~Vandergheynst.
\newblock {Convolutional Neural Networks on Graphs with Fast Localized Spectral
  Filtering}.
\newblock In {\em {Proceedings of the 30th International Conference on Neural
  Information Processing Systems (NIPS' 16)}}, pp. 3844--3852, 2016.
  \href{https://doi.org/10.5555/3157382.3157527}
{doi: \textsf{%
10\hspace{.1pt}\discretionary{.}{%
}{.}\hspace{.4pt}5555\discretionary{/}{%
}{/}3157382\hspace{.1pt}\discretionary{.}{%
}{.}\hspace{.4pt}3157527}}


\bibitem{dengSurveyUrbanVisual2023}
Z.~Deng, D.~Weng, S.~Liu, Y.~Tian, M.~Xu, and Y.~Wu.
\newblock {A Survey of Urban Visual Analytics: {{Advances}} and Future
  Directions}.
\newblock {\em Comp. Visual Media}, 9(1):3--39, 2023.
  \href{https://doi.org/10.1007/s41095-022-0275-7}
{doi: \textsf{%
10\hspace{.1pt}\discretionary{.}{%
}{.}\hspace{.4pt}1007\discretionary{/}{%
}{/}s41095\discretionary{%
}{-}{-}022\discretionary{%
}{-}{-}0275\discretionary{%
}{-}{-}7}}


\bibitem{fasshauerChoosingOptimalShape2007}
G.~E. Fasshauer and J.~G. Zhang.
\newblock {On Choosing ``Optimal'' Shape Parameters for {{RBF}} Approximation}.
\newblock {\em Numer Algor}, 45(1-4):345--368, 2007.
  \href{https://doi.org/10.1007/s11075-007-9072-8}
{doi: \textsf{%
10\hspace{.1pt}\discretionary{.}{%
}{.}\hspace{.4pt}1007\discretionary{/}{%
}{/}s11075\discretionary{%
}{-}{-}007\discretionary{%
}{-}{-}9072\discretionary{%
}{-}{-}8}}


\bibitem{fengTopologyDensityMap2021}
Z.~Feng, H.~Li, W.~Zeng, S.-H. Yang, and H.~Qu.
\newblock {Topology {{Density Map}} for {{Urban Data Visualization}} and
  {{Analysis}}}.
\newblock {\em IEEE Transactions on Visualization and Computer Graphics},
  27(2):828--838, 2021. \href{https://doi.org/10.1109/TVCG.2020.3030469}
{doi: \textsf{%
10\hspace{.1pt}\discretionary{.}{%
}{.}\hspace{.4pt}1109\discretionary{/}{%
}{/}TVCG\hspace{.1pt}\discretionary{.}{%
}{.}\hspace{.4pt}2020\hspace{.1pt}\discretionary{.}{%
}{.}\hspace{.4pt}3030469}}


\bibitem{gershonVisualizationImperfectWorld1998}
N.~Gershon.
\newblock {Visualization of an Imperfect World}.
\newblock {\em IEEE Computer Graphics and Applications}, 18(4):43--45, 1998.
  \href{https://doi.org/10.1109/38.689662}
{doi: \textsf{%
10\hspace{.1pt}\discretionary{.}{%
}{.}\hspace{.4pt}1109\discretionary{/}{%
}{/}38\hspace{.1pt}\discretionary{.}{%
}{.}\hspace{.4pt}689662}}


\bibitem{hamiltonInductiveRepresentationLearning2017}
W.~L. Hamilton, R.~Ying, and J.~Leskovec.
\newblock {Inductive Representation Learning on Large Graphs}.
\newblock In {\em {Proceedings of the 30th International Conference on Neural
  Information Processing Systems (NIPS' 17)}}, {{NIPS}}'17, 2017.

\bibitem{henglRegressionkrigingEquationsCase2007}
T.~Hengl, G.~B.~M. Heuvelink, and D.~G. Rossiter.
\newblock {About Regression-Kriging: {{From}} Equations to Case Studies}.
\newblock {\em Computers \& Geosciences}, 33(10):1301--1315, 2007.
  \href{https://doi.org/10.1016/j.cageo.2007.05.001}
{doi: \textsf{%
10\hspace{.1pt}\discretionary{.}{%
}{.}\hspace{.4pt}1016\discretionary{/}{%
}{/}j\hspace{.1pt}\discretionary{.}{%
}{.}\hspace{.4pt}cageo\hspace{.1pt}\discretionary{.}{%
}{.}\hspace{.4pt}2007\hspace{.1pt}\discretionary{.}{%
}{.}\hspace{.4pt}05\hspace{.1pt}\discretionary{.}{%
}{.}\hspace{.4pt}001}}


\bibitem{huberRobustEstimationLocation1964}
P.~J. Huber.
\newblock {Robust {{Estimation}} of a {{Location Parameter}}}.
\newblock {\em Breakthroughs in Statistics}, pp. 492--518, 1964.
  \href{https://doi.org/10.1007/978-1-4612-4380-9_35}
{doi: \textsf{%
10\hspace{.1pt}\discretionary{.}{%
}{.}\hspace{.4pt}1007\discretionary{/}{%
}{/}978\discretionary{%
}{-}{-}1\discretionary{%
}{-}{-}4612\discretionary{%
}{-}{-}4380\discretionary{%
}{-}{-}9\_35}}


\bibitem{hullmanPursuitErrorSurvey2019}
J.~Hullman, X.~Qiao, M.~Correll, A.~Kale, and M.~Kay.
\newblock {In {{Pursuit}} of {{Error}}: {{A Survey}} of {{Uncertainty
  Visualization Evaluation}}}.
\newblock {\em IEEE Transactions on Visualization and Computer Graphics},
  25(1):903--913, 2019. \href{https://doi.org/10.1109/TVCG.2018.2864889}
{doi: \textsf{%
10\hspace{.1pt}\discretionary{.}{%
}{.}\hspace{.4pt}1109\discretionary{/}{%
}{/}TVCG\hspace{.1pt}\discretionary{.}{%
}{.}\hspace{.4pt}2018\hspace{.1pt}\discretionary{.}{%
}{.}\hspace{.4pt}2864889}}


\bibitem{jainStructuralRNNDeepLearning2016}
A.~Jain, A.~R. Zamir, S.~Savarese, and A.~Saxena.
\newblock {Structural-{{RNN}}: {{Deep Learning}} on {{Spatio-Temporal
  Graphs}}}.
\newblock {\em 2016 IEEE Conference on Computer Vision and Pattern Recognition
  (CVPR)}, pp. 5308--5317, 2016. \href{https://doi.org/10.1109/CVPR.2016.573}
{doi: \textsf{%
10\hspace{.1pt}\discretionary{.}{%
}{.}\hspace{.4pt}1109\discretionary{/}{%
}{/}CVPR\hspace{.1pt}\discretionary{.}{%
}{.}\hspace{.4pt}2016\hspace{.1pt}\discretionary{.}{%
}{.}\hspace{.4pt}573}}


\bibitem{joslynCommunicatingForecastUncertainty2010}
S.~Joslyn and S.~Savelli.
\newblock {Communicating Forecast Uncertainty: Public Perception of Weather
  Forecast Uncertainty}.
\newblock {\em Meteorological Applications}, 17(2):180--195, 2010.
  \href{https://doi.org/10.1002/met.190}
{doi: \textsf{%
10\hspace{.1pt}\discretionary{.}{%
}{.}\hspace{.4pt}1002\discretionary{/}{%
}{/}met\hspace{.1pt}\discretionary{.}{%
}{.}\hspace{.4pt}190}}


\bibitem{kayWhenIshMy2016}
M.~Kay, T.~Kola, J.~R. Hullman, and S.~A. Munson.
\newblock {When (Ish) Is {{My Bus}}?: {{User-centered Visualizations}} of
  {{Uncertainty}} in {{Everyday}}, {{Mobile Predictive Systems}}}.
\newblock In {\em {Proceedings of the 2016 {{CHI Conference}} on {{Human
  Factors}} in {{Computing Systems}}}}, pp. 5092--5103. San Jose, 2016.
  \href{https://doi.org/10.1145/2858036.2858558}
{doi: \textsf{%
10\hspace{.1pt}\discretionary{.}{%
}{.}\hspace{.4pt}1145\discretionary{/}{%
}{/}2858036\hspace{.1pt}\discretionary{.}{%
}{.}\hspace{.4pt}2858558}}


\bibitem{kingmaAdamMethodStochastic2014}
D.~P. Kingma and J.~Ba.
\newblock {Adam: {{A Method}} for {{Stochastic Optimization}}}.
\newblock In {\em 3rd International Conference on Learning Representations,
  {ICLR} 2015}, 2015.

\bibitem{kinkeldeyHowAssessVisual2014}
C.~Kinkeldey, A.~M. MacEachren, and J.~Schiewe.
\newblock {How to {{Assess Visual Communication}} of {{Uncertainty}}? {{A
  Systematic Review}} of {{Geospatial Uncertainty Visualisation User
  Studies}}}.
\newblock {\em The Cartographic Journal}, 51(4):372--386, 2014.
  \href{https://doi.org/10.1179/1743277414Y.0000000099}
{doi: \textsf{%
10\hspace{.1pt}\discretionary{.}{%
}{.}\hspace{.4pt}1179\discretionary{/}{%
}{/}1743277414Y\hspace{.1pt}\discretionary{.}{%
}{.}\hspace{.4pt}0000000099}}


\bibitem{kirkwoodBayesianDeepLearning2022}
C.~Kirkwood, T.~Economou, N.~Pugeault, and H.~Odbert.
\newblock {Bayesian {{Deep Learning}} for {{Spatial Interpolation}} in the
  {{Presence}} of {{Auxiliary Information}}}.
\newblock {\em Math Geosci}, 54(3):507--531, 2022.
  \href{https://doi.org/10.1007/s11004-021-09988-0}
{doi: \textsf{%
10\hspace{.1pt}\discretionary{.}{%
}{.}\hspace{.4pt}1007\discretionary{/}{%
}{/}s11004\discretionary{%
}{-}{-}021\discretionary{%
}{-}{-}09988\discretionary{%
}{-}{-}0}}


\bibitem{klemmer2023positional}
K.~Klemmer, N.~S. Safir, and D.~B. Neill.
\newblock {Positional Encoder Graph Neural Networks for Geographic Data}.
\newblock In {\em {Proceedings of the 26th International Conference on
  Artificial Intelligence and Statistics (AISTATS) 2023}}, pp. 1379--1389.

\bibitem{klippelInterpretingSpatialPatterns2011}
A.~Klippel, F.~Hardisty, and R.~Li.
\newblock {Interpreting {{Spatial Patterns}}: {{An Inquiry Into Formal}} and
  {{Cognitive Aspects}} of {{Tobler}}'s {{First Law}} of {{Geography}}}.
\newblock {\em Annals of the Association of American Geographers},
  101(5):1011--1031, 2011. \href{https://doi.org/10.1080/00045608.2011.577364}
{doi: \textsf{%
10\hspace{.1pt}\discretionary{.}{%
}{.}\hspace{.4pt}1080\discretionary{/}{%
}{/}00045608\hspace{.1pt}\discretionary{.}{%
}{.}\hspace{.4pt}2011\hspace{.1pt}\discretionary{.}{%
}{.}\hspace{.4pt}577364}}


\bibitem{korporaalEffectsUncertaintyVisualization2020}
M.~Korporaal, I.~T. Ruginski, and S.~I. Fabrikant.
\newblock {Effects of {{Uncertainty {{Visualization}}}} on {{Map-Based Decision
  Making Under Time Pressure}}}.
\newblock {\em Frontiers in Computer Science}, 2, 2020.
  \href{https://doi.org/10.3389/fcomp.2020.00032}
{doi: \textsf{%
10\hspace{.1pt}\discretionary{.}{%
}{.}\hspace{.4pt}3389\discretionary{/}{%
}{/}fcomp\hspace{.1pt}\discretionary{.}{%
}{.}\hspace{.4pt}2020\hspace{.1pt}\discretionary{.}{%
}{.}\hspace{.4pt}00032}}


\bibitem{lamSpatialInterpolationMethods1983}
N.~S.-N. Lam.
\newblock {Spatial {{Interpolation Methods}}: {{A Review}}}.
\newblock {\em The American Cartographer}, 10(2):129--150, 1983.
  \href{https://doi.org/10.1559/152304083783914958}
{doi: \textsf{%
10\hspace{.1pt}\discretionary{.}{%
}{.}\hspace{.4pt}1559\discretionary{/}{%
}{/}152304083783914958}}


\bibitem{laslettComparisonSeveralSpatial1987}
G.~M. Laslett, A.~B. McBratney, P.~J. Pahl, and M.~F. Hutchinson.
\newblock {Comparison of Several Spatial Prediction Methods for Soil {{pH}}}.
\newblock {\em Journal of Soil Science}, 38(2):325--341, 1987.
  \href{https://doi.org/10.1111/j.1365-2389.1987.tb02148.x}
{doi: \textsf{%
10\hspace{.1pt}\discretionary{.}{%
}{.}\hspace{.4pt}1111\discretionary{/}{%
}{/}j\hspace{.1pt}\discretionary{.}{%
}{.}\hspace{.4pt}1365\discretionary{%
}{-}{-}2389\hspace{.1pt}\discretionary{.}{%
}{.}\hspace{.4pt}1987\hspace{.1pt}\discretionary{.}{%
}{.}\hspace{.4pt}tb02148\hspace{.1pt}\discretionary{.}{%
}{.}\hspace{.4pt}x}}


\bibitem{liStreamMapSmoothDynamic2018}
C.~Li, G.~Baciu, and Y.~Han.
\newblock {{{StreamMap}}: {{Smooth Dynamic Visualization}} of {{High-Density
  Streaming Points}}}.
\newblock {\em IEEE Transactions on Visualization and Computer Graphics},
  24(3):1381--1393, 2018. \href{https://doi.org/10.1109/TVCG.2017.2668409}
{doi: \textsf{%
10\hspace{.1pt}\discretionary{.}{%
}{.}\hspace{.4pt}1109\discretionary{/}{%
}{/}TVCG\hspace{.1pt}\discretionary{.}{%
}{.}\hspace{.4pt}2017\hspace{.1pt}\discretionary{.}{%
}{.}\hspace{.4pt}2668409}}


\bibitem{liSpatialInterpolationMethods2014}
J.~Li and A.~D. Heap.
\newblock {Spatial Interpolation Methods Applied in the Environmental Sciences:
  {{A}} Review}.
\newblock {\em Environmental Modelling \& Software}, 53:173--189, 2014.
  \href{https://doi.org/10.1016/j.envsoft.2013.12.008}
{doi: \textsf{%
10\hspace{.1pt}\discretionary{.}{%
}{.}\hspace{.4pt}1016\discretionary{/}{%
}{/}j\hspace{.1pt}\discretionary{.}{%
}{.}\hspace{.4pt}envsoft\hspace{.1pt}\discretionary{.}{%
}{.}\hspace{.4pt}2013\hspace{.1pt}\discretionary{.}{%
}{.}\hspace{.4pt}12\hspace{.1pt}\discretionary{.}{%
}{.}\hspace{.4pt}008}}


\bibitem{liDiffusionConvolutionalRecurrent2018}
Y.~Li, R.~Yu, C.~Shahabi, and Y.~Liu.
\newblock {Diffusion {{Convolutional Recurrent Neural Network}}: {{Data-Driven
  Traffic Forecasting}}}.
\newblock 2018.

\bibitem{liuTPFlowProgressivePartition2019}
D.~Liu, P.~Xu, and L.~Ren.
\newblock {{{TPFlow}}: {{Progressive Partition}} and {{Multidimensional Pattern
  Extraction}} for {{Large-Scale Spatio-Temporal Data Analysis}}}.
\newblock {\em IEEE Transactions on Visualization and Computer Graphics},
  25(1):1--11, 2019. \href{https://doi.org/10.1109/TVCG.2018.2865018}
{doi: \textsf{%
10\hspace{.1pt}\discretionary{.}{%
}{.}\hspace{.4pt}1109\discretionary{/}{%
}{/}TVCG\hspace{.1pt}\discretionary{.}{%
}{.}\hspace{.4pt}2018\hspace{.1pt}\discretionary{.}{%
}{.}\hspace{.4pt}2865018}}


\bibitem{maceachrenVisualizingGeospatialInformation2005}
A.~M. MacEachren, A.~Robinson, S.~Hopper, S.~Gardner, R.~Murray, M.~Gahegan,
  and E.~Hetzler.
\newblock {Visualizing {{Geospatial Information Uncertainty}}: {{What We Know}}
  and {{What We Need}} to {{Know}}}.
\newblock {\em Cartography and Geographic Information Science}, 32(3):139--160,
  2005. \href{https://doi.org/10.1559/1523040054738936}
{doi: \textsf{%
10\hspace{.1pt}\discretionary{.}{%
}{.}\hspace{.4pt}1559\discretionary{/}{%
}{/}1523040054738936}}


\bibitem{maiMultiScaleRepresentationLearning2020}
G.~Mai, K.~Janowicz, B.~Yan, R.~Zhu, L.~Cai, and N.~Lao.
\newblock {Multi-{{Scale Representation Learning}} for {{Spatial Feature
  Distributions}} Using {{Grid Cells}}}, 2020.

\bibitem{matheronPrinciplesGeostatistics1963}
G.~Matheron.
\newblock {Principles of Geostatistics}.
\newblock {\em Economic Geology}, 58(8):1246--1266, 1963.
  \href{https://doi.org/10.2113/gsecongeo.58.8.1246}
{doi: \textsf{%
10\hspace{.1pt}\discretionary{.}{%
}{.}\hspace{.4pt}2113\discretionary{/}{%
}{/}gsecongeo\hspace{.1pt}\discretionary{.}{%
}{.}\hspace{.4pt}58\hspace{.1pt}\discretionary{.}{%
}{.}\hspace{.4pt}8\hspace{.1pt}\discretionary{.}{%
}{.}\hspace{.4pt}1246}}


\bibitem{menneUnitedStatesHistorical2012}
M.~Menne and C.~Williams.
\newblock {United {{States Historical Climatology Network}} ({{USHCN}}),
  {{Version}} 2.5.5.20231225}, 2012. \href{https://doi.org/10.7289/V56W98B4}
{doi: \textsf{%
10\hspace{.1pt}\discretionary{.}{%
}{.}\hspace{.4pt}7289\discretionary{/}{%
}{/}V56W98B4}}


\bibitem{miaoExperimentalSurveyMissing2023}
X.~Miao, Y.~Wu, L.~Chen, Y.~Gao, and J.~Yin.
\newblock {An {{Experimental Survey}} of {{Missing Data Imputation
  Algorithms}}}.
\newblock {\em IEEE Transactions on Knowledge and Data Engineering},
  35(7):6630--6650, 2023. \href{https://doi.org/10.1109/TKDE.2022.3186498}
{doi: \textsf{%
10\hspace{.1pt}\discretionary{.}{%
}{.}\hspace{.4pt}1109\discretionary{/}{%
}{/}TKDE\hspace{.1pt}\discretionary{.}{%
}{.}\hspace{.4pt}2022\hspace{.1pt}\discretionary{.}{%
}{.}\hspace{.4pt}3186498}}


\bibitem{mitasSpatialInterpolation1999}
L.~Mitas and H.~Mitasova.
\newblock {Spatial {{Interpolation}}}.
\newblock In {\em {Geographical {{Information Systems}}: {{Principles}},
  {{Techniques}}, {{Management}} and {{Applications}}, 2nd {{Edition}},
  {{Abridged}}}}, pp. 481--492. 1999.

\bibitem{myersSpatialInterpolationOverview1994}
D.~E. Myers.
\newblock {Spatial Interpolation: An Overview}.
\newblock {\em Geoderma}, 62(1):17--28, 1994.
  \href{https://doi.org/10.1016/0016-7061(94)90025-6}
{doi: \textsf{%
10\hspace{.1pt}\discretionary{.}{%
}{.}\hspace{.4pt}1016\discretionary{/}{%
}{/}0016\discretionary{%
}{-}{-}7061\discretionary{%
}{(}{(}94\discretionary{)}{%
}{)}90025\discretionary{%
}{-}{-}6}}


\bibitem{nowakDesigningAmbiguityVisual2023}
S.~Nowak and L.~Bartram.
\newblock {Designing for {{Ambiguity}} in {{Visual Analytics}}: {{Lessons}}
  from {{Risk Assessment}} and {{Prediction}}}.
\newblock {\em IEEE Trans. Visual. Comput. Graphics}, pp. 1--10, 2023.
  \href{https://doi.org/10.1109/TVCG.2023.3326571}
{doi: \textsf{%
10\hspace{.1pt}\discretionary{.}{%
}{.}\hspace{.4pt}1109\discretionary{/}{%
}{/}TVCG\hspace{.1pt}\discretionary{.}{%
}{.}\hspace{.4pt}2023\hspace{.1pt}\discretionary{.}{%
}{.}\hspace{.4pt}3326571}}


\bibitem{oliverKrigingMethodInterpolation1990}
M.~A. OLIVER and R.~WEBSTER.
\newblock {Kriging: A Method of Interpolation for Geographical Information
  Systems}.
\newblock {\em International Journal of Geographical Information Systems},
  4(3):313--332, 1990. \href{https://doi.org/10.1080/02693799008941549}
{doi: \textsf{%
10\hspace{.1pt}\discretionary{.}{%
}{.}\hspace{.4pt}1080\discretionary{/}{%
}{/}02693799008941549}}


\bibitem{pangApproachesUncertaintyVisualization1997}
A.~T. Pang, C.~M. Wittenbrink, and S.~K. Lodha.
\newblock {Approaches to Uncertainty Visualization}.
\newblock {\em The Visual Computer}, 13(8):370--390, 1997.
  \href{https://doi.org/10.1007/s003710050111}
{doi: \textsf{%
10\hspace{.1pt}\discretionary{.}{%
}{.}\hspace{.4pt}1007\discretionary{/}{%
}{/}s003710050111}}


\bibitem{prestonCommunicatingUncertaintyRisk2023}
A.~Preston and K.-L. Ma.
\newblock {Communicating {{Uncertainty}} and {{Risk}} in {{Air Quality Maps}}}.
\newblock {\em IEEE Transactions on Visualization and Computer Graphics},
  29(9):3746--3757, 2023. \href{https://doi.org/10.1109/TVCG.2022.3171443}
{doi: \textsf{%
10\hspace{.1pt}\discretionary{.}{%
}{.}\hspace{.4pt}1109\discretionary{/}{%
}{/}TVCG\hspace{.1pt}\discretionary{.}{%
}{.}\hspace{.4pt}2022\hspace{.1pt}\discretionary{.}{%
}{.}\hspace{.4pt}3171443}}


\bibitem{retchlessGuidanceRepresentingUncertainty2016}
D.~P. Retchless and C.~A. Brewer.
\newblock {Guidance for Representing Uncertainty on Global Temperature Change
  Maps}.
\newblock {\em International Journal of Climatology}, 36(3):1143--1159, 2016.
  \href{https://doi.org/10.1002/joc.4408}
{doi: \textsf{%
10\hspace{.1pt}\discretionary{.}{%
}{.}\hspace{.4pt}1002\discretionary{/}{%
}{/}joc\hspace{.1pt}\discretionary{.}{%
}{.}\hspace{.4pt}4408}}


\bibitem{sarmaEvaluatingUseUncertainty2023}
A.~Sarma, S.~Guo, J.~Hoffswell, R.~Rossi, F.~Du, E.~Koh, and M.~Kay.
\newblock {Evaluating the {{Use}} of {{Uncertainty Visualisations}} for
  {{Imputations}} of {{Data Missing At Random}} in {{Scatterplots}}}.
\newblock {\em IEEE Transactions on Visualization and Computer Graphics},
  29(1):602--612, 2023. \href{https://doi.org/10.1109/TVCG.2022.3209348}
{doi: \textsf{%
10\hspace{.1pt}\discretionary{.}{%
}{.}\hspace{.4pt}1109\discretionary{/}{%
}{/}TVCG\hspace{.1pt}\discretionary{.}{%
}{.}\hspace{.4pt}2022\hspace{.1pt}\discretionary{.}{%
}{.}\hspace{.4pt}3209348}}


\bibitem{seoStructuredSequenceModeling2018}
Y.~Seo, M.~Defferrard, P.~Vandergheynst, and X.~Bresson.
\newblock {Structured {{Sequence Modeling}} with {{Graph Convolutional
  Recurrent Networks}}}.
\newblock In L.~Cheng, A.~C.~S. Leung, and S.~Ozawa, eds., {\em {Proceedings of
  the 31th International Conference on Neural Information Processing Systems
  (NIPS' 18)}}, vol. 11301, pp. 362--373. Cham, 2018.
  \href{https://doi.org/10.1007/978-3-030-04167-0_33}
{doi: \textsf{%
10\hspace{.1pt}\discretionary{.}{%
}{.}\hspace{.4pt}1007\discretionary{/}{%
}{/}978\discretionary{%
}{-}{-}3\discretionary{%
}{-}{-}030\discretionary{%
}{-}{-}04167\discretionary{%
}{-}{-}0\_33}}


\bibitem{silvermanDensityEstimizationStatistics1986}
B.~W. Silverman.
\newblock {\em {Density {{Estimization}} for {{Statistics}} and {{Data
  Analysis}}}}.
\newblock London, United Kingdom, 1986.

\bibitem{smiejaProcessingMissingData2018}
M.~{\'S}mieja, L.~Struski, J.~Tabor, B.~Zieli{\'n}ski, and P.~Spurek.
\newblock {Processing of Missing Data by Neural Networks}.
\newblock In {\em {Proceedings of the 32th International Conference on Neural
  Information Processing Systems (NIPS' 18)}}, 2018.

\bibitem{toblerComputerMovieSimulating1970}
W.~R. Tobler.
\newblock {A {{Computer Movie Simulating Urban Growth}} in the {{Detroit
  Region}}}.
\newblock {\em Economic Geography}, 46:234, 1970.
  \href{https://doi.org/10.2307/143141}
{doi: \textsf{%
10\hspace{.1pt}\discretionary{.}{%
}{.}\hspace{.4pt}2307\discretionary{/}{%
}{/}143141}}


\bibitem{unitedstatesenvironmentalprotectionagencyepaAirDataAir2016}
{United States Environmental Protection Agency (EPA)}.
\newblock {Air {{Data}}: {{Air Quality Data Collected}} at {{Outdoor Monitors
  Across}} the {{US}}}.
\newblock https://www.epa.gov/outdoor-air-quality-data, 2016.

\bibitem{vaswaniAttentionAllYou2017}
A.~Vaswani, N.~Shazeer, N.~Parmar, J.~Uszkoreit, L.~Jones, A.~N. Gomez, L.~u.
  Kaiser, and I.~Polosukhin.
\newblock Attention is all you need.
\newblock In {\em Advances in Neural Information Processing Systems}, vol.~30,
  2017.

\bibitem{velickovicGraphAttentionNetworks2018}
P.~Veli{\v c}kovi{\'c}, G.~Cucurull, A.~Casanova, A.~Romero, P.~Li{\`o}, and
  Y.~Bengio.
\newblock {Graph {{Attention Networks}}}, 2018.
  \href{https://doi.org/10.48550/arXiv.1710.10903}
{doi: \textsf{%
10\hspace{.1pt}\discretionary{.}{%
}{.}\hspace{.4pt}48550\discretionary{/}{%
}{/}arXiv\hspace{.1pt}\discretionary{.}{%
}{.}\hspace{.4pt}1710\hspace{.1pt}\discretionary{.}{%
}{.}\hspace{.4pt}10903}}


\bibitem{wuInductiveGraphNeural2020}
Y.~Wu, D.~Zhuang, A.~Labbe, and L.~Sun.
\newblock {Inductive Graph Neural Networks for Spatiotemporal Kriging}.
\newblock {\em Proceedings of the AAAI Conference on Artificial Intelligence},
  35(5):4478--4485, 2021. \href{https://doi.org/10.1609/aaai.v35i5.16575}
{doi: \textsf{%
10\hspace{.1pt}\discretionary{.}{%
}{.}\hspace{.4pt}1609\discretionary{/}{%
}{/}aaai\hspace{.1pt}\discretionary{.}{%
}{.}\hspace{.4pt}v35i5\hspace{.1pt}\discretionary{.}{%
}{.}\hspace{.4pt}16575}}


\bibitem{wuSpatialAggregationTemporal2021}
Y.~Wu, D.~Zhuang, M.~Lei, A.~Labbe, and L.~Sun.
\newblock {Spatial {{Aggregation}} and {{Temporal Convolution Networks}} for
  {{Real-time Kriging}}}, 2021.
  \href{https://doi.org/10.48550/arXiv.2109.12144}
{doi: \textsf{%
10\hspace{.1pt}\discretionary{.}{%
}{.}\hspace{.4pt}48550\discretionary{/}{%
}{/}arXiv\hspace{.1pt}\discretionary{.}{%
}{.}\hspace{.4pt}2109\hspace{.1pt}\discretionary{.}{%
}{.}\hspace{.4pt}12144}}


\bibitem{wuComprehensiveSurveyGraph2021}
Z.~Wu, S.~Pan, F.~Chen, G.~Long, C.~Zhang, and P.~S. Yu.
\newblock {A {{Comprehensive Survey}} on {{Graph Neural Networks}}}.
\newblock {\em IEEE Transactions on Neural Networks and Learning Systems},
  32(1):4--24, 2021. \href{https://doi.org/10.1109/TNNLS.2020.2978386}
{doi: \textsf{%
10\hspace{.1pt}\discretionary{.}{%
}{.}\hspace{.4pt}1109\discretionary{/}{%
}{/}TNNLS\hspace{.1pt}\discretionary{.}{%
}{.}\hspace{.4pt}2020\hspace{.1pt}\discretionary{.}{%
}{.}\hspace{.4pt}2978386}}


\bibitem{xu2020STMFGCN}
Z.~Xu, Y.~Kang, Y.~Cao, and Z.~Li.
\newblock {Spatiotemporal Graph Convolution Multifusion Network for Urban
  Vehicle Emission Prediction}.
\newblock 32(8):3342--3354. \href{https://doi.org/10.1109/TNNLS.2020.3008702}
{doi: \textsf{%
10\hspace{.1pt}\discretionary{.}{%
}{.}\hspace{.4pt}1109\discretionary{/}{%
}{/}TNNLS\hspace{.1pt}\discretionary{.}{%
}{.}\hspace{.4pt}2020\hspace{.1pt}\discretionary{.}{%
}{.}\hspace{.4pt}3008702}}


\bibitem{yanSurveyBlueNoiseSampling2015}
D.-M. Yan, J.-W. Guo, B.~Wang, X.-P. Zhang, and P.~Wonka.
\newblock {A {{Survey}} of {{Blue-Noise Sampling}} and {{Its Applications}}}.
\newblock {\em J. Comput. Sci. Technol.}, 30(3):439--452, 2015.
  \href{https://doi.org/10.1007/s11390-015-1535-0}
{doi: \textsf{%
10\hspace{.1pt}\discretionary{.}{%
}{.}\hspace{.4pt}1007\discretionary{/}{%
}{/}s11390\discretionary{%
}{-}{-}015\discretionary{%
}{-}{-}1535\discretionary{%
}{-}{-}0}}


\bibitem{yanHighAccuracyInterpolation2021}
L.~Yan, X.~Tang, and Y.~Zhang.
\newblock {High {{Accuracy Interpolation}} of {{DEM Using Generative
  Adversarial Network}}}.
\newblock {\em Remote Sensing}, 13(4):676, 2021.
  \href{https://doi.org/10.3390/rs13040676}
{doi: \textsf{%
10\hspace{.1pt}\discretionary{.}{%
}{.}\hspace{.4pt}3390\discretionary{/}{%
}{/}rs13040676}}


\bibitem{yanSpatialTemporalGraph2018}
S.~Yan, Y.~Xiong, and D.~Lin.
\newblock {Spatial {{Temporal Graph Convolutional Networks}} for
  {{Skeleton-Based Action Recognition}}}.
\newblock In {\em {{{Proceedings of the AAAI Conference on Artificial
  Intelligence}}}}, vol.~32, 2018.
  \href{https://doi.org/10.1609/aaai.v32i1.12328}
{doi: \textsf{%
10\hspace{.1pt}\discretionary{.}{%
}{.}\hspace{.4pt}1609\discretionary{/}{%
}{/}aaai\hspace{.1pt}\discretionary{.}{%
}{.}\hspace{.4pt}v32i1\hspace{.1pt}\discretionary{.}{%
}{.}\hspace{.4pt}12328}}


\bibitem{yaoSpatiotemporalInterpolationUsing2023}
S.~Yao and B.~Huang.
\newblock {Spatiotemporal {{Interpolation Using Graph Neural Network}}}.
\newblock {\em Annals of the American Association of Geographers},
  113(8):1856--1877, 2023. \href{https://doi.org/10.1080/24694452.2023.2206469}
{doi: \textsf{%
10\hspace{.1pt}\discretionary{.}{%
}{.}\hspace{.4pt}1080\discretionary{/}{%
}{/}24694452\hspace{.1pt}\discretionary{.}{%
}{.}\hspace{.4pt}2023\hspace{.1pt}\discretionary{.}{%
}{.}\hspace{.4pt}2206469}}


\bibitem{yinGPS2VecGeneratingWorldwide2019}
Y.~Yin, Z.~Liu, Y.~Zhang, S.~Wang, R.~R. Shah, and R.~Zimmermann.
\newblock {{{GPS2Vec}}: {{Towards Generating Worldwide GPS Embeddings}}}.
\newblock In {\em {Proceedings of the 27th {{ACM SIGSPATIAL International
  Conference}} on {{Advances}} in {{Geographic Information Systems}}}}, pp.
  416--419. Chicago IL USA, 2019.
  \href{https://doi.org/10.1145/3347146.3359067}
{doi: \textsf{%
10\hspace{.1pt}\discretionary{.}{%
}{.}\hspace{.4pt}1145\discretionary{/}{%
}{/}3347146\hspace{.1pt}\discretionary{.}{%
}{.}\hspace{.4pt}3359067}}


\bibitem{yoonGAINMissingData2018}
J.~Yoon, J.~Jordon, and M.~Schaar.
\newblock {{{GAIN}}: {{Missing Data Imputation}} Using {{Generative Adversarial
  Nets}}}.
\newblock In {\em 2018 International Conference of Machine Learning}, 2018.

\bibitem{yuSpatioTemporalGraphConvolutional2018}
B.~Yu, H.~Yin, and Z.~Zhu.
\newblock {Spatio-{{Temporal Graph Convolutional Networks}}: {{A Deep Learning
  Framework}} for {{Traffic Forecasting}}}.
\newblock In {\em {Proceedings of the {{Twenty-Seventh International Joint
  Conference}} on {{Artificial Intelligence}}}}, pp. 3634--3640. Stockholm,
  Sweden, 2018. \href{https://doi.org/10.24963/ijcai.2018/505}
{doi: \textsf{%
10\hspace{.1pt}\discretionary{.}{%
}{.}\hspace{.4pt}24963\discretionary{/}{%
}{/}ijcai\hspace{.1pt}\discretionary{.}{%
}{.}\hspace{.4pt}2018\discretionary{/}{%
}{/}505}}


\bibitem{zhangUncertaintyOrientedEnsembleData2021}
M.~Zhang, L.~Chen, Q.~Li, X.~Yuan, and J.~Yong.
\newblock {Uncertainty-{{Oriented Ensemble Data Visualization}} and
  {{Exploration}} Using {{Variable Spatial Spreading}}}.
\newblock {\em IEEE Transactions on Visualization and Computer Graphics},
  27(2):1808--1818, 2021. \href{https://doi.org/10.1109/TVCG.2020.3030377}
{doi: \textsf{%
10\hspace{.1pt}\discretionary{.}{%
}{.}\hspace{.4pt}1109\discretionary{/}{%
}{/}TVCG\hspace{.1pt}\discretionary{.}{%
}{.}\hspace{.4pt}2020\hspace{.1pt}\discretionary{.}{%
}{.}\hspace{.4pt}3030377}}


\bibitem{zhangIntegrationMachineLearning2021}
S.~E. Zhang, G.~T. Nwaila, L.~Tolmay, H.~E. Frimmel, and J.~E. Bourdeau.
\newblock {Integration of {{Machine Learning Algorithms}} with {{Gompertz
  Curves}} and {{Kriging}} to {{Estimate Resources}} in {{Gold Deposits}}}.
\newblock {\em Nat Resour Res}, 30(1):39--56, 2021.
  \href{https://doi.org/10.1007/s11053-020-09750-z}
{doi: \textsf{%
10\hspace{.1pt}\discretionary{.}{%
}{.}\hspace{.4pt}1007\discretionary{/}{%
}{/}s11053\discretionary{%
}{-}{-}020\discretionary{%
}{-}{-}09750\discretionary{%
}{-}{-}z}}


\bibitem{zhouVoidsFillingMultiattention2022}
G.~Zhou, B.~Song, P.~Liang, J.~Xu, and T.~Yue.
\newblock {Voids {{Filling}} of {{DEM}} with {{Multiattention Generative
  Adversarial Network Model}}}.
\newblock {\em Remote Sensing}, 14(5):1206, 2022.
  \href{https://doi.org/10.3390/rs14051206}
{doi: \textsf{%
10\hspace{.1pt}\discretionary{.}{%
}{.}\hspace{.4pt}3390\discretionary{/}{%
}{/}rs14051206}}


\bibitem{zhuSpatialInterpolationUsing2020}
D.~Zhu, X.~Cheng, F.~Zhang, X.~Yao, Y.~Gao, and Y.~Liu.
\newblock {Spatial Interpolation Using Conditional Generative Adversarial
  Neural Networks}.
\newblock {\em International Journal of Geographical Information Science},
  34(4):735--758, 2020. \href{https://doi.org/10.1080/13658816.2019.1599122}
{doi: \textsf{%
10\hspace{.1pt}\discretionary{.}{%
}{.}\hspace{.4pt}1080\discretionary{/}{%
}{/}13658816\hspace{.1pt}\discretionary{.}{%
}{.}\hspace{.4pt}2019\hspace{.1pt}\discretionary{.}{%
}{.}\hspace{.4pt}1599122}}


\bibitem{zhuInferringSpatialInteraction2018}
D.~Zhu, Z.~Huang, L.~Shi, L.~Wu, and Y.~Liu.
\newblock {Inferring Spatial Interaction Patterns from Sequential Snapshots of
  Spatial Distributions}.
\newblock {\em International Journal of Geographical Information Science},
  32(4):783--805, 2018. \href{https://doi.org/10.1080/13658816.2017.1413192}
{doi: \textsf{%
10\hspace{.1pt}\discretionary{.}{%
}{.}\hspace{.4pt}1080\discretionary{/}{%
}{/}13658816\hspace{.1pt}\discretionary{.}{%
}{.}\hspace{.4pt}2017\hspace{.1pt}\discretionary{.}{%
}{.}\hspace{.4pt}1413192}}


\end{thebibliography}
\end{document}